\documentclass[10pt,twocolumn,letterpaper]{article}

\usepackage{wacv}              
\usepackage{multirow}

\definecolor{myred}{RGB}{230, 121, 138}
\definecolor{mygreen}{RGB}{106, 175, 70}
\definecolor{myblue}{RGB}{89, 161, 230}

\definecolor{paired}{RGB}{111, 148, 230}
\definecolor{unpaired}{RGB}{234, 134, 119}

\newcommand{\method}{GenD\xspace}
\newcommand{\PEcoreL}{$\text{PE}_\text{core}\text{L}$\xspace}

\def\myparagraph#1{\medskip\noindent{\bf{#1}}}

\usepackage{graphicx}
\usepackage{makecell}
\usepackage{booktabs}
\usepackage{enumitem}

\definecolor{wacvblue}{rgb}{0.21,0.49,0.74}
\usepackage[pagebackref,breaklinks,colorlinks,allcolors=wacvblue]{hyperref}

\def\wacvPaperID{1353} 
\def\confName{WACV}
\def\confYear{2026}

\title{Deepfake Detection that Generalizes Across Benchmarks}

\author{
Andrii Yermakov\textsuperscript{1} \quad
Jan Cech\textsuperscript{1} \quad
Jiri Matas\textsuperscript{1} \quad
Mario Fritz\textsuperscript{2} \\
\textsuperscript{1}Czech Technical University in Prague \quad
\textsuperscript{2}CISPA Helmholtz Center for Information Security \\
{\tt\small \{yermaand,cechj,matas\}@fel.cvut.cz} \quad
{\tt\small fritz@cispa.de}
}

\begin{document}
\maketitle
\begin{abstract}

The generalization of deepfake detectors to unseen manipulation techniques remains a challenge for practical deployment. Although many approaches adapt foundation models by introducing significant architectural complexity, this work demonstrates that robust generalization is achievable through a parameter-efficient adaptation of one of the foundational pre-trained vision encoders. The proposed method, \method, fine-tunes only the Layer Normalization parameters (0.03\% of the total) and enhances generalization by enforcing a hyperspherical feature manifold using L2 normalization and metric learning on it.

We conducted an extensive evaluation on 14 benchmark datasets spanning from 2019 to 2025. The proposed method achieves state-of-the-art performance, outperforming more complex, recent approaches in average cross-dataset AUROC. Our analysis yields two primary findings for the field: 1) training on paired real-fake data from the same source video is essential for mitigating shortcut learning and improving generalization, and 2) detection difficulty on academic datasets has not strictly increased over time, with models trained on older, diverse datasets showing strong generalization capabilities. 

This work delivers a computationally efficient and reproducible method, proving that state-of-the-art generalization is attainable by making targeted, minimal changes to a pre-trained foundational image encoder model. The code is at: \url{https://github.com/yermandy/GenD}.

\end{abstract}    
\section{Introduction}
\label{sec:introduction}

The proliferation of realistic facial deepfakes raises significant concerns regarding misinformation and malicious use, with AI-manipulated videos -- those altered by techniques like face swapping or face reenactment -- making detection challenging. Unlike fully synthetic content, such forgeries preserve the original context and leave subtle artifacts that are difficult for humans and machines to detect~\cite{human-study-1, human-study-2}. 

A primary issue affecting current detection methods is their limited ability to generalize. A model that has been trained to identify images altered by a particular deepfake generation algorithm often struggles when faced with examples produced by a new generation algorithm. 

The generalization gap is the primary issue that we address in this work. Assuming the hypothesis that \textit{adapted} large-scale, pre-trained foundational vision encoder can serve as a general foundation for deepfake detection~\cite{UniFD}, we build the proposed method in three variants, using Contrastive Language-Image Pre-training (CLIP)~\cite{CLIP}, Perception Encoder (PE)~\cite{PE} and DINO~\cite{DINOv3} models as feature extractors, which are known for their generalizable visual representations. 

The proposed method consists of a vision encoder, whose outputs are L2-normalized. We then fine-tune only the parameters of the Layer Normalization blocks~\cite{LN-tuning} while keeping the rest frozen. Additionally, we propose using metric learning in this L2 space to enhance generalization. 

We benchmarked the generalization capabilities of the proposed model on 14 deepfake video datasets released between 2019 and 2025, listed in ~\cref{tab:test-datasets}. To our knowledge, this represents the broadest evaluation in the deepfake literature. We show that the proposed model outperforms the most recent state-of-the-art methods on the majority of all available benchmarks.

In summary, our \textbf{key contributions} are as follows:

\begin{itemize}
    \item A novel deepfake detection method called \method. The method achieves the best average cross-dataset AUROC compared to recently released models.

    \item The most comprehensive evaluation in the deepfake literature covering datasets released throughout six years of research.
    
    \item A demonstration that to achieve the best generalization and prevent shortcut learning, it is essential to construct the training set consisting of real-fake pairs, where the fake video is generated from the real counterpart of the pair.
\end{itemize}

\section{Related Work}

Deepfake techniques have rapidly evolved. Key methods include: Face-swap~\cite{faceswap}, Face-reenactment~\cite{face-reanectment}, LipSync~\cite{lipsync}, and full face video synthesis~\cite{synthesia, veo3}. The first three approaches manipulate only small, localized areas of the video, typically the facial region, while leaving the rest of the footage untouched, making detection challenging. The techniques are continually improving, leveraging advanced models to produce increasingly seamless and realistic results without obvious, visually perceptible artifacts.

As deepfake production techniques advance, detection methods are also evolving to keep pace. The literature encompasses a taxonomy of detection strategies, which can be divided into two main branches: content-based and signal-based approaches. Content-based methods focus on visible or interpretable inconsistencies, utilizing inductive biases such as apparent blending boundaries~\cite{SBI}, unsynchronized audio-visual movements~\cite{LipForensics}, or motion artifacts~\cite{agarwal2019protecting}. Signal-based methods, on the other hand, detect subtle and often invisible traces left in the visual signal, fingerprints that reveal synthetic origins even when no overt artifacts are present. Signal-based techniques have become increasingly accurate, especially as newer deepfakes leave fewer perceptible cues. However, a persistent challenge remains: generalizing to previously unseen manipulation methods. 

Recent efforts in detecting general AI-generated content~\cite{UniFD, CLIPping, SSL-in-DF, FatFormer} and facial deepfakes specifically~\cite{UDD, LipFD, DFD-FCG, ForensicsAdapter, Effort} have begun to adopt CLIP as a backbone for generalizable deepfake detection.
These approaches can be broadly categorized by the type of model adaptation. Some methods, like \textbf{Forensics Adapter}~\cite{ForensicsAdapter} (ForAda), introduce \textit{architectural} changes by adding a separate parallel network explicitly trained to identify artifacts left by blending. Other methods operate on the \textit{parameter space}; for instance, \textbf{Effort}~\cite{Effort} uses Singular Value Decomposition (SVD) to decompose the model's weights into orthogonal subspaces, freezing the principal components to preserve pre-trained knowledge while fine-tuning the residual components on forgery patterns.

In contrast to these approaches, our work proposes to modify a subset of parameters. LN-Tuning~\cite{LN-tuning} adjusts the affine parameters of Layer Normalization blocks. The efficiency of LN-tuning in the general context was analyzed in~\cite{LN-tuning-1, LN-tuning-2}. We demonstrate that it achieves competitive or state-of-the-art performance in the deepfake detection task.

As a complimentary approach to the passive detection methods of AI manipulations, there exist active defense approaches that protect real images by incorporating invisible watermarks~\cite{zhao2024proactive, asnani2023malp, bartolucci2025perturb, zhai2023defending, asnani2022proactive, yang2021faceguard}.

\section{Method}

\begin{figure}[t]
    \centering
    \includegraphics[width=1.0\linewidth]{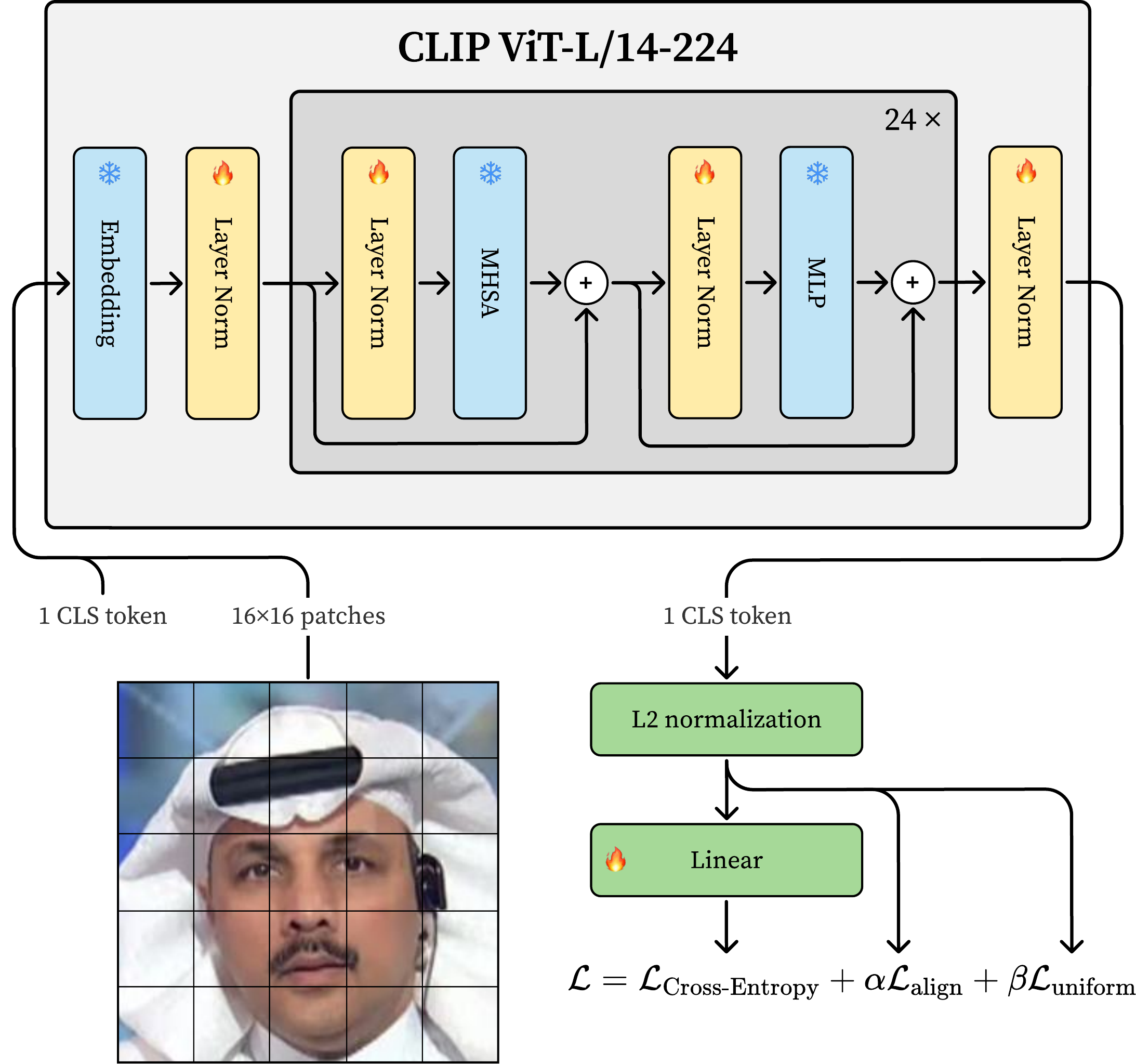}
    \caption{\label{fig:model}
        The architecture of \method (CLIP). The gray rectangle represents the original CLIP ViT image encoder, and green represents the added components. Only rectangles with a fire icon represent layers with trainable parameters.}
\end{figure}

The proposed method comes in three variants depending on the pre-trained image encoder to which we add two components: L2 normalization of the classification token and a linear classifier. We optimize only 0.03\% of network parameters using a weighted combination of cross-entropy, uniformity, and alignment losses in an end-to-end fashion. We call this approach \method, and its complete overview is shown in \cref{fig:model}.

\subsection{Model}
\label{sec:model}

\myparagraph{Feature extractor.}
We use three different pre-trained image encoders: 1. 
CLIP ViT-L/14, trained by OpenAI~\cite{CLIP}; 2. \PEcoreL, trained by Meta AI~\cite{PE}; 3. DINOv3 ViT-L/16~\cite{DINOv3}, trained by Meta AI. After the encoder processes the image, we take only the $1024$-dimensional classification token and discard the rest.


\myparagraph{Optimized parameters.}
The L2-normalized classification token is used as the input to the linear binary classifier, which gives logits for fake and real classes. In addition to the weights $W\in\mathbb{R}^{1024\times2}$ and biases $b\in\mathbb{R}^2$ of the classification layer, we optimize the parameters of all Layer Normalization blocks~\cite{LN-tuning, LN-tuning-1, LN-tuning-2}. This roughly constitutes 0.03\% of the total model parameters; the remaining parameters of the encoder are frozen.

\myparagraph{Loss function.}
The network is trained using a weighted combination of cross-entropy, uniformity, and alignment losses~\cite{UniAlign}
\begin{equation} \label{eq:loss}
\mathcal{L} = \mathcal{L}_{\text{Cross-Entropy}} + \alpha \mathcal{L}_{\text{align}} + \beta \mathcal{L}_{\text{uniform}}.
\end{equation}
The cross-entropy loss is applied to fake and real logits, while uniformity and alignment losses are directly applied to the L2-normalized classification token, see~\cref{fig:model}.

Let $\mathbf{z}_i$ be the L2-normalized feature of the $i$-th sample. The alignment loss tightens feature clusters coming from the same class
\begin{equation} \label{eq:align}
\mathcal{L}_{\text{align}} = \mathop{\mathbb{E}}\limits_{x, y \sim \mathcal{P}^+}\left[\|\mathbf{z}_x - \mathbf{z}_{y}\|_2^2\right],
\end{equation}
\noindent where $\mathcal{P}^+$ is the distribution of positive pairs. Uniformity encourages the features to be uniformly distributed on the unit hypersphere

\begin{equation} \label{eq:uniform}
\mathcal{L}_{\text{uniform}} = \log \mathop{\mathbb{E}}\limits_{x, y \sim \mathcal{P}}\left[e^{-2\|\mathbf{z}_x - \mathbf{z}_y\|_2^2}\right],
\end{equation}

\noindent where $x$ and $y$ are sampled independently from the data distribution $\mathcal{P}$.

\myparagraph{Training algorithm.}
To optimize the model parameters, we use Adam optimizer~\cite{Adam} with $\beta_1=0.9$ and $\beta_2=0.999$ without weight decay. The weight precision is set to bfloat16. The learning rate is scheduled using a cosine cyclic rule~\cite{cyclic}. Each cycle starts with a linear warm-up for one epoch, increasing from $10^{-5}$ to $3\times10^{-4}$, and then decays using cosine scheduling over nine epochs to $10^{-5}$. For most runs, we observed that the training did not improve after two cycles (or 20 epochs). The batch size is set to 128 samples. For CLIP ViT-L/14, the loss coefficients of \cref{eq:loss} are set to $\alpha=0.1$ and $\beta=0.5$, found on the validation set.

\myparagraph{Video classification.}
\method has a single frame as an input. To get video-level probabilities, we uniformly sample 32 frames from a video, calculate softmax probabilities for every frame independently, and average them.

\myparagraph{Speed and memory.}
Every encoder of \method is of the size of a ViT-L model, consisting of roughly 304M parameters. Our non-optimized for speed and memory implementation in PyTorch can process 120 frames per second on A100 with a batch size of 1, requiring approximately 2GB of VRAM.

\subsection{Data}

\myparagraph{Data preprocessing.}
We follow a standard practice of dataset preprocessing proposed in DeepfakeBench~\cite{DeepfakeBench}, which consists of the following steps:

\begin{enumerate}
  \item Sample 32 frames evenly from each video.
  \item Extract the largest face using RetinaFace~\cite{RetinaFace}.
  \item Align the face using predicted landmarks.
  \item Calculate the bounding box for the aligned face.
  \item Enlarge the bounding box by a $1.3\times$ margin.
  \item Crop and resize the face to $224\times224$.
  \item Save the image in lossless format.
\end{enumerate}

\myparagraph{Training data.}
Following the standard protocol, we train the model on the FaceForensics++ dataset~\cite{FF++} (FF++), which consists of $3600$ videos, of which $720$ are real videos, and the rest, $4\times720$, come from four different deepfake forgery methods. After preprocessing, we obtain approximately $115$k frames. The dataset was released in three different compression levels. Following all of the preceding work, we use the c23 compression version of the dataset. We augment the training set using image augmentations such as random horizontal flipping, random affine transformations, Gaussian blurring, color jitter, and JPEG compression, all implemented in the \texttt{torchvision} package. In ablations, we also experimented with changing the training set to FFIW~\cite{FFIW} and DSv2~\cite{DSv1}.

\myparagraph{Training samples pairing.}
We train the model on a dataset consisting of real-fake pairs, where each fake video is generated from its real counterpart. We hypothesize and empirically confirm in~\cref{sec:real_fake_pairs} that the dataset constructed in this way forces the network to learn the subtle, low-level artifacts of the manipulation process, rather than exploiting superficial, high-level differences between unrelated videos, thus preventing shortcut learning. 

\myparagraph{Validation data.}
We noticed that the FF++ validation set is very similar to the training set. That makes the comparison of different training experiments on such a validation ineffective. Each experiment achieves the best validation AUROC quickly, shows almost no signs of overfitting, and does not provide any estimate of cross-dataset generalization. The FF++ validation set does not provide crucial information on how the methods will perform when trained for an extended period. Therefore, we create a custom validation set, which consists of data from validation or training splits of other datasets, namely CDFv3~\cite{CDFv3}, DSv1 and DSv2~\cite{DSv1}, and FFIW~\cite{FFIW}. We ensure that test samples from these datasets are not present in the custom validation set. 

\section{Experiments}
\label{sec:experiments}

\begin{table}[t]
\centering
    \captionof{table}{Test datasets. The number of real and fake videos in the datasets. Negative values: the number of videos missing in our experiments, compared to the original datasets, due to face detector failures. Values in rows with datasets marked $*$ represent the number of videos randomly subsampled from the original dataset. Gen. means the number of generators used in the dataset.
    }
    \label{tab:test-datasets}
\tabcolsep=4.2pt
\begin{tabular}{@{}rlcrlrl@{}}
\toprule
\textbf{Year}  & \textbf{Dataset}  &\textbf{Gen.}& \multicolumn{1}{r}{\textbf{Real}} &  & \multicolumn{1}{r}{\textbf{Fake}} &  \\ \midrule
2019  & FF++~\cite{FF++}  &4& 140 &  & 560 &  \\
2019  & DF~\cite{FF++}  &1& 140 &  & 140 &  \\
2019  & F2F~\cite{FF++}  &1& 140 &  & 140 &  \\
2019  & FS~\cite{FF++}  &1& 140 &  & 140 &  \\
2019  & NT~\cite{FF++}  &1& 140 &  & 140 &  \\
2019  & DFD~\cite{DFD}  &5& 363 &  & 3068 & -2 \\
2019  & UADFV~\cite{UADFV}  &1& 49 &  & 49 &  \\
2019  & DFDC~\cite{DFDC}  &8& 2500 & -185 & 2500 & -111 \\
2020  & FSh~\cite{FSh}  &1& 140 &  & 140 &  \\
2020  & CDFv2~\cite{CDFv2}  &1& 178 &  & 340 &  \\
2021  & FFIW~\cite{FFIW}  &3& 1738 & -3 & 1738 & -3 \\
2021  & KoDF~\cite{KoDF} $*$  &6& 403 &  & 1106 &  \\
2021  & FAVC~\cite{FakeAVCeleb}  &4& 500 &  & 20566 & -22 \\
2022  & DFDM~\cite{DFDM}  &5& 590 & -2 & 1720 & -2 \\
2024  & PGF~\cite{PolyGlotFake}  &10& 762 &  & 13605 &  \\
2024  & IDF~\cite{IDForge} $*$  &9& 18834 &  & 2323 &  \\
2024  & DSv1.1~\cite{DSv1}  &5& 1416 &  & 1497 &  \\
2025  & DSv2~\cite{DSv1}  &6& 1863 &  & 1416 &  \\
2025  & CDFv3~\cite{CDFv3}  &22& 178 &  & 5240 & -1  \\ \bottomrule
\end{tabular}
\end{table}


We demonstrate the generalizability of \method by presenting a series of experiments. First, we introduce the diverse range of benchmark datasets and the evaluation metric. Next, we conduct a comprehensive cross-dataset evaluation, where we train the model on the FF++ dataset and compare its performance against numerous state-of-the-art (SOTA) methods on other benchmarks. Following this, a detailed ablation study is presented to systematically analyze the contribution of each component of the proposed method. We then investigated two hypotheses: one demonstrating the critical importance of training on paired real-fake data for robust generalization and another exploring the evolution of deepfake detection difficulty over the years, which shows that training on older, diverse datasets can be more beneficial than using solely recent datasets. We finish the experiments with robustness evaluation against common image degradation techniques.

We ensure the statistical significance of the results by repeating each experiment 5 times with different training seeds. Every reported AUROC for \method except for the \cref{tab:methods-comparison} is averaged over 5 seeds; standard deviations can be found in the supplementary text.

\subsection{Test benchmarks}

For reporting the performance of the model in cross-dataset settings, we are using Celeb-DF-v2 (CDFv2)~\cite{CDFv2}, Celeb-DF++ (CDFv3)~\cite{CDFv3}, DeepFake Detection Challenge (DFDC)~\cite{DFDC}, Google's DFD dataset~\cite{DFD}, Face Forensics in the Wild (FFIW)~\cite{FFIW}, DeepSpeak v1.1 (DSv1.1) and DeepSpeak v2.0 (DSv2)~\cite{DSv1}, Korean DeepFake Detection Dataset (KoDF)~\cite{KoDF}, FakeAVCeleb (FAVC)~\cite{FakeAVCeleb}, DeepFakes from Different Models (DFDM)~\cite{DFDM}, PolyGlotFake (PGF)~\cite{PolyGlotFake}, and IDForge (IDF)~\cite{IDForge}. The statistics for the test part of the datasets are shown in ~\cref{tab:test-datasets}.

We are using the video-level area under the ROC curve (AUROC) as the main comparison and optimization metric. See the supplementary material for more evaluation metrics such as average precision (AP) or equal error rate (EER).

\subsection{Cross-dataset evaluation}

\begin{table*}[t]
\centering
\caption{\label{tab:methods-comparison}
Video-level AUROC (\%) in cross-dataset testing of models trained on the FF++ dataset. Results of other methods are taken from their original papers. Values with superscript citations are taken from the papers referenced in superscripts. Video input means that the model takes a sequence of frames and models temporal correlations between them.}

\begin{tabular}{lccclcccc}
\toprule
\textbf{Model}  & \textbf{Year} & \textbf{Publication}  & \textbf{Input}  &\textbf{Backbone} & \textbf{CDFv2} & \textbf{DFD} & \textbf{DFDC}  & \textbf{FFIW}   \\
\midrule
LipForensics~\cite{LipForensics}     & 2021 & CVPR& Video &ResNet-18& 82.4           & --             & 73.5  &   --     \\
FTCN~\cite{FTCN}                     & 2021 & ICCV   & Video &3D ResNet-50& 86.9           & --             & 74.0  & $74.5^\text{\cite{SBI}}$\\
RealForensics~\cite{RealForensics}   & 2022 & CVPR   & Video &Modified CSN& 86.9           & --             & 75.9  &   --       \\
SBI~\cite{SBI}                       & 2022 & CVPR   & Frame &EFNB4& 93.2& 82.7& 72.4&  84.8\\
AUNet~\cite{AUNet}                   & 2023 & CVPR   & Video &Xception+ART& 92.8& \textbf{99.2}& 73.8&  81.5\\
StyleDFD~\cite{StyleDFD}             & 2024 & CVPR   & Video &3D ResNet-50& 89.0           & 96.1           & --    &   --       \\
LSDA~\cite{LSDA}                     & 2024 & CVPR   & Frame &EFNB4& 91.1           & --             & 77.0  & $72.4^{\text{\cite{Effort}}}$   \\
LAA-Net~\cite{LAA-Net}               & 2024 & CVPR   & Frame &EFNB4& 95.4           & 98.4& 86.9&   --       \\
AltFreezing~\cite{AltFreezing}       & 2024 & CVPR   & Video &3D ResNet-50& 89.5           & 98.5  & --  &   --       \\
NACO~\cite{NACO}                     & 2024 & ECCV   & Video &ViT-B/16& 89.5           & --             & 76.7  &   --       \\
 RAE~\cite{RAE} & 2024& ECCV & Frame &ViT-B/16& 95.5& 99.0& 80.2&--\\
TALL++~\cite{TALL++}                 & 2024 & IJCV   & Video &Swin-B& 92.0& --             & 78.5&   --       \\
 ProDet~\cite{ProDet} & 2024& NeurIPS & Frame &EFNB4& 92.5& --             & 77.0& --              \\
UDD~\cite{UDD}                       & 2025 &  AAAI & Frame &CLIP ViT-B/16& 93.1& 95.5& 81.2&   --       \\
 P\&P~\cite{plug-and-play} & 2025& CVPR & Video &CLIP ViT-L/14& 94.7& 96.5& 84.3&92.1\\
 DFD-FCG~\cite{DFD-FCG} & 2025& CVPR & Video &CLIP ViT-L/14& 95.0& --& 81.8&--\\
 ForAda~\cite{ForensicsAdapter}   & 2025 &  CVPR & Frame &CLIP ViT-L/14& 95.7     & 97.2           & \textbf{87.2}  & --   \\
 Effort~\cite{Effort}                 & 2025 &  ICML & Frame &CLIP ViT-L/14& 95.6           & 96.5           & 84.3  & 92.1   \\
\midrule
\method (CLIP) & 2025 &  -- & Frame & CLIP ViT-L/14 & \textbf{96.0} & 97.0 & 87.1 & 92.8 \\
\method (PE) & 2025 &  -- & Frame & \PEcoreL & 95.8 & 96.5 & 81.6 & 93.3 \\
 \method (DINO) & 2025 &  -- & Frame & DINOv3 ViT-L/16 & 92.2 & 96.6 & 84.7 & \textbf{94.5} \\
 \bottomrule
\end{tabular}
\end{table*}

We compared the proposed model in a cross-dataset fashion following the standard protocol, that is, by training on FF++ and testing on the rest of the datasets as in previous work. \cref{tab:methods-comparison} compares the video-level AUROC of the proposed model with SOTA approaches. AUROC values for the competing methods were taken directly from the original papers. The model achieves SOTA results on two datasets: CDFv2 and FFIW, and gets high results on the DFDC dataset.

The proposed method achieves these results simply by averaging individual frame softmax probabilities, without complex post-processing or architectural modifications as described in \cref{sec:model}. This contrasts with many contemporary methods that rely on complex multimodal analyses, such as modeling temporal inconsistencies between video frames or detecting audio-video discrepancies~\cite{LipForensics, LipFD, RealForensics, FTCN, DFD-FCG, plug-and-play}. The performance of the proposed method suggests that the features of the visual encoder, when properly tuned, are highly effective for this detection task.

We also evaluated the generalization of the proposed method against current SOTA models across a wide range of benchmark datasets. For this experiment, we selected the most recent open-source SOTA models that have made their code and pre-trained weights publicly available, namely ForAda~\cite{ForensicsAdapter} and Effort~\cite{Effort}. To ensure that each model was used as intended by its original authors, we utilized their officially released weights and integrated their data preprocessing pipelines which were used during model training. After reproducing the results on the test datasets that were used in these papers, we obtained the same or very similar results, presented in \cref{tab:comparison-reported-reproduced}. The in-distribution results on FF++ are shown in \cref{tab:in-dataset-auroc}.

Subsequently, we compared the pre-trained models with the proposed model on the comprehensive collection of test datasets from \cref{tab:test-datasets}. The results of this large-scale evaluation are presented in \cref{tab:reproduced-cross-dataset}. This methodology ensures that the methods were compared on the same data. From the results of this experiment, it is seen that the proposed approach generalized better, which is proved by higher average performance across all datasets.

\begin{table}[t]
    \centering
    \caption{
    \label{tab:in-dataset-auroc}
    Video-level test AUROC (\%) on in-domain FF++ dataset calculated by us. \method results are the means over five seeds.}
    \begin{tabular}{l|cccc|c}
        \toprule
         \textbf{Method} & \textbf{DF} & \textbf{F2F} & \textbf{FS} & \textbf{NT} & \textbf{Mean}\\
        \midrule
ForAda~\cite{ForensicsAdapter} & \textbf{99.7} & 97.0 & 98.6 & 91.9 & 96.8 \\
Effort~\cite{Effort} & 99.4 & 93.2 & 98.4 & 84.6 & 93.9\\
GenD (CLIP) & 99.5 & 98.1 & 98.7 & 95.5 & 98.0 \\
GenD (PE) & \textbf{99.7} & 99.3 & 99.1 & \textbf{97.5} & \textbf{98.9} \\
GenD (DINO) & 99.6 & \textbf{99.4} & \textbf{99.4} & 96.7 & 98.8 \\

        \bottomrule
    \end{tabular}
\end{table}

\begin{table*}[t]
    \centering
    \caption{
    \label{tab:reproduced-cross-dataset}
    Cross-dataset video-level AUROC (\%) for reproduced methods. The highest score in each column is in bold. Results for \method are the averages over five training seeds; standard deviations can be found in the supplementary.}

    \resizebox{\textwidth}{!}{%
    \begin{tabular}{l|cccccccccccccc|c}
        \toprule
         \textbf{Method} & \thead{2019 \\ \textbf{UADFV}} & \thead{2019 \\ \textbf{DFD}} & \thead{2019 \\ \textbf{DFDC}} & \thead{2020 \\ \textbf{FSh}} & \thead{2020 \\ \textbf{CDFv2}} & \thead{2021 \\ \textbf{FFIW}} & \thead{2021 \\ \textbf{KoDF}} & \thead{2021 \\ \textbf{FAVC}} & \thead{2022 \\ \textbf{DFDM}} & \thead{2024 \\ \textbf{PGF}} & \thead{2024 \\ \textbf{IDF}} & \thead{2024 \\ \textbf{DSv1.1}} & \thead{2025 \\ \textbf{DSv2}}  &\thead{2025  \\ \textbf{CDFv3}}& \textbf{Mean} \\
        \midrule
ForAda~\cite{ForensicsAdapter} & \textbf{99.4} & \textbf{97.2} & \textbf{87.3} & 82.0 & \textbf{95.7} & 90.6 & 88.2 & 93.1 & 97.1 & 86.6 & 90.8 & 81.8 & 72.8 & 75.6 & 88.4 \\
Effort~\cite{Effort} & 97.4 & 95.2 & 85.4 & \textbf{91.2} & 93.2 & 92.5 & 88.1 & 92.4 & 98.2 & 84.9 & 96.0 & 82.1 & 64.4 & 78.7 & 88.5 \\
GenD (CLIP) & 99.2 & 96.4 & 86.4 & 86.6 & 94.6 & 91.5 & 84.9 & 96.0 & 99.6 & 89.6 & 97.8 & \textbf{90.1} & 77.7 & 85.9 & 91.2 \\
GenD (PE) & 97.7 & 96.8 & 82.2 & 87.6 & 95.0 & \textbf{93.7} & 85.1 & 97.3 & 98.3 & 92.3 & 97.9 & 87.8 & 78.6 & \textbf{89.5} & 91.4 \\
GenD (DINO) & 98.6 & 96.2 & 85.6 & 88.8 & 92.5 & 92.9 & \textbf{89.7} & \textbf{98.4} & \textbf{99.8} & \textbf{92.4} & \textbf{98.2} & 86.9 & \textbf{79.4} & 83.5 & \textbf{91.6} \\

        \bottomrule
    \end{tabular}%
    }
\end{table*}

\begin{table}
\centering
\caption{Comparison of Reported / Reproduced video-level AUROC (\%) for reimplemented deepfake detectors.}
\label{tab:comparison-reported-reproduced}
\tabcolsep=3.5pt
\begin{tabular}{@{}l r@{\,/\,}l r@{\,/\,}l r@{\,/\,}l r@{\,/\,}l@{}}
\toprule
\textbf{Method} & \multicolumn{2}{c}{\textbf{CDFv2}} & \multicolumn{2}{c}{\textbf{DFD}} & \multicolumn{2}{c}{\textbf{DFDC}} & \multicolumn{2}{c}{\textbf{FFIW}} \\
\midrule
ForAda~\cite{ForensicsAdapter} & 95.7 & 95.7 & 97.2 & 97.2 & 87.2 & 87.3 & --   & 90.6 \\
Effort~\cite{Effort} & 95.6 & 93.2 & 96.5 & 95.2 & 84.3 & 85.4 & 92.1 & 92.5 \\
\bottomrule
\end{tabular}
\end{table}

\subsection{Ablation studies}

We conducted ablation experiments to understand the contribution of each component; see \cref{tab:ablation}. There are three setups tested for every backbone. Setup 3 is the proposed \method:

\begin{enumerate}
    \item \textbf{Baseline}: the setup is similar to~\cite{UniFD}. We freeze the backbone and train a linear classifier with cross-entropy loss on top of the feature space of the classification token.
    \item \textbf{Baseline + LN}: adds tuning of Layer Normalization parameters to setup 1.
    \item \textbf{Baseline + LN + UA}: L2-normalizes classification token and adds uniformity and alignment losses to setup 2.
\end{enumerate}

\begin{table*}[ht]
    \centering
    \caption{
    \label{tab:ablation}
        Ablation study. Impact of model components on test video-level AUROC in cross-dataset settings. Each setup was trained on the FF++ training set. Reported values are averages over 5 training seeds. Setups \#3, \#6, \#9 are the proposed \method.
    }
    \tabcolsep=4pt
    \resizebox{\textwidth}{!}{
    \begin{tabular}{rl|cccccccccccccc|c}
        \toprule
        \textbf{\#} & \textbf{Setup}  &\textbf{UADFV}& \textbf{DFD}  & \textbf{DFDC}    &\textbf{FSh}&\textbf{CDFv2} & \textbf{FFIW}  &\textbf{KoDF} &\textbf{FAVC} &\textbf{DFDM} &\textbf{PGF} &\textbf{IDF}& \textbf{DSv1.1}&\textbf{DSv2}&\textbf{CDFv3} &\textbf{Mean}\\
        \midrule
1&CLIP + LP & 94.6 & 89.2 & 75.3 & 77.6 & 74.6 & 80.7 & 81.8 & 83.0 & 77.8 & 62.7 & 68.7 & 64.5 & 57.8 & 75.6 & 76.0 \\
2&\#1 + LN & 99.0 & 96.1 & 85.4 & \textbf{86.9} & 93.8 & \textbf{91.5} & 84.4 & 95.5 & 99.3 & 87.5 & 95.0 & 89.1 & 74.0 & \textbf{86.8} & 90.3 \\
3&\#2 + UA & \textbf{99.2} & \textbf{96.4} & \textbf{86.4} & 86.6 & \textbf{94.6} & 91.5 & \textbf{84.9} & \textbf{96.0} & \textbf{99.6} & \textbf{89.6} & \textbf{97.8} & \textbf{90.1} & \textbf{77.7} & 85.9 & \textbf{91.2} \\
\midrule

4&\PEcoreL + LP & 76.1 & 76.4 & 67.2 & 80.3 & 72.2 & 75.7 & 65.6 & 71.3 & 73.3 & 52.8 & 74.6 & 65.6 & 59.5 & 79.5 & 70.7 \\
5&\#4 + LN & \textbf{98.4} & 96.4 & 81.9 & \textbf{88.1} & 94.2 & 93.0 & 84.5 & 96.5 & 97.9 & 91.2 & 97.7 & \textbf{88.0} & 77.0 & \textbf{89.9} & 91.0 \\
6&\#5 + UA & 97.7 & \textbf{96.8} & \textbf{82.2} & 87.6 & \textbf{95.0} & \textbf{93.7} & \textbf{85.1} & \textbf{97.3} & \textbf{98.3} & \textbf{92.3} & \textbf{97.9} & 87.8 & \textbf{78.6} & 89.5 & \textbf{91.4} \\
\midrule

7&DINOv3L + LP & 84.2 & 79.8 & 69.0 & 71.9 & 68.2 & 74.7 & 75.4 & 81.1 & 85.9 & 65.0 & 82.0 & 72.2 & 70.2 & 68.3 & 74.9 \\
8&\#7 + LN & 97.3 & 94.8 & 84.4 & 88.7 & 91.7 & 90.4 & \textbf{90.6} & 98.4 & 98.8 & \textbf{93.6} & 97.5 & \textbf{89.1} & \textbf{82.4} & 80.1 & 91.3 \\
9&\#8 + UA & \textbf{98.6} & \textbf{96.2} & \textbf{85.6} & \textbf{88.8} & \textbf{92.5} & \textbf{92.9} & 89.7 & \textbf{98.4} & \textbf{99.8} & 92.4 & \textbf{98.2} & 86.9 & 79.4 & \textbf{83.5} & \textbf{91.6} \\

\bottomrule
    \end{tabular}
    }
    
\end{table*}

\myparagraph{Layer Normalization tuning.}
As shown in \cref{tab:ablation}, LN-tuning consistently achieves substantial gains in AUROC across all benchmarks.

We explore further parameter-efficient fine-tuning (PEFT) strategies, the goal of which is to modify only a relatively small number of parameters while keeping most parameters untouched, thus efficiently introducing several degrees of freedom. PEFT is more effective for generalization than full fine-tuning (FFT) in low-data regimes~\cite{PEFT}, which is particularly applicable to deepfake detection, where collected datasets and benchmarks have a limited number of samples. This work explores three PEFT methods: LoRA~\cite{hu2022lora}, LN-tuning~\cite{LN-tuning, CLIPFit}, and bias tuning (BitFit)~\cite{CLIPFit, BitFit}. 

We found that FFT leads to rapid overfitting and poor generalization across datasets. This observation is consistent with recent findings from Effort~\cite{Effort} that FFT degrades the generalization performance. 

Similarly to~\cite{Effort}, we fine-tuned the parameters of every transformer weight matrix using LoRA to preserve the generalization. Setting LoRA to rank one, the lowest possible, allowed the model to reach a near-perfect training AUROC (99.99\%) in two epochs, resulting in rapid overfitting. 

We observed that BitFit~\cite{BitFit} and LN-tuning~\cite{LN-tuning} have stable training dynamics by first improving the validation AUROC, reaching a peak, and overfitting after some epochs. In comparison, FFT and LoRA show no improvement in validation AUROC even after a single training epoch.

Experimental results showed that in combination with other components, LN-tuning achieves the best performance and prevents overfitting compared to other PEFT strategies, such as BitFit, LoRA, or FFT. These results suggest that LN tuning works as a good tuning strategy for this task. A more detailed ablation comparing these techniques can be found in the supplementary material. 

\myparagraph{Uniformity and Alignment.}
We hypothesized that representations that effectively exploit the latent space could help us improve generalization. We used uniformity and alignment loss~\cite{UniAlign}, leading to slower overfitting and better validation AUROC. We suppose that adding these losses helps to utilize the hyperspherical space more effectively, resulting in improved generalization.

\subsection{Importance of training on paired dataset}
\label{sec:real_fake_pairs}

A fundamental question in training deepfake detectors is how to construct training data more effectively to promote generalization. In this section, we empirically validate the hypothesis that achieving better generalization requires \textit{a training dataset in which each fake video has a real counterpart from which it was produced}.

\begin{figure}[t]
    \centering
   \begin{subfigure}[b]{0.225\textwidth}
        \centering
        \caption{\textcolor{paired}{Paired}}
        \includegraphics[width=\linewidth]{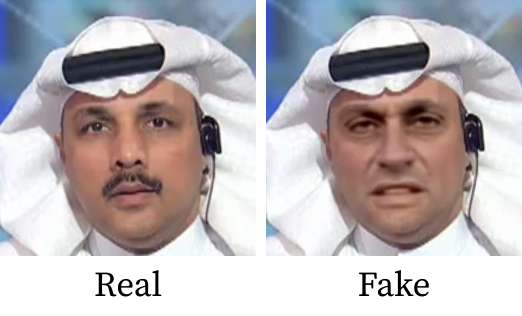}
    \end{subfigure}
    \hfill
    \begin{subfigure}[b]{0.225\textwidth}
        \centering
        \caption{\textcolor{unpaired}{Unpaired}}
        \includegraphics[width=\linewidth]{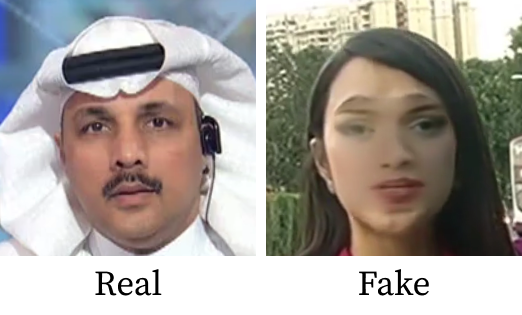}
    \end{subfigure}  
    \caption{
        Samples from (a) \textcolor{paired}{Paired} and (b) \textcolor{unpaired}{Unpaired} datasets.
    }
  
    \label{fig:paired_unpaired_faces}
\end{figure}

To empirically validate this hypothesis, we designed an experiment with two differently constructed training datasets. The first dataset, denoted as \textcolor{paired}{paired}, consists of data where for each fake video there exists its pair -- a real counterpart from which this fake was created. In the second dataset, \textcolor{unpaired}{unpaired}, fake videos never have its pair -- a real counterpart. Examples of samples from these two datasets can be found in \cref{fig:paired_unpaired_faces}.

In particular, we generate 20 different paired and 20 unpaired dataset splits from the FF++ training set. Every split shares the real part but has a different fake part. In the paired dataset, every fake video shares the background with a video from the real part. In an unpaired dataset, every fake video does not have a corresponding real video with the same background, see \cref{fig:paired_unpaired_faces}. We ensure that the number of fake and real videos between the paired and unpaired datasets is the same.

\begin{figure}[t]
    \centering
     \begin{subfigure}[b]{0.235\textwidth}
        \centering
        \caption{Training}
        \includegraphics[width=\linewidth]{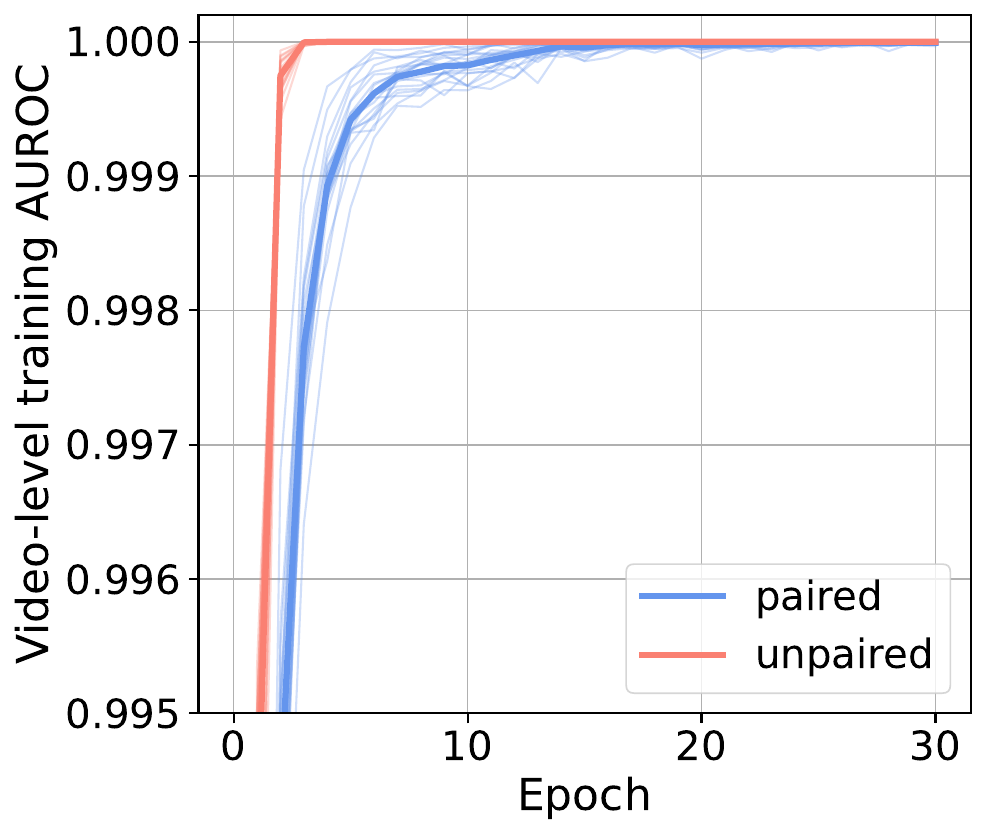}
    \end{subfigure}
    \hfill
    \begin{subfigure}[b]{0.235\textwidth}
        \centering
        \caption{Validation}
        \includegraphics[width=\linewidth]{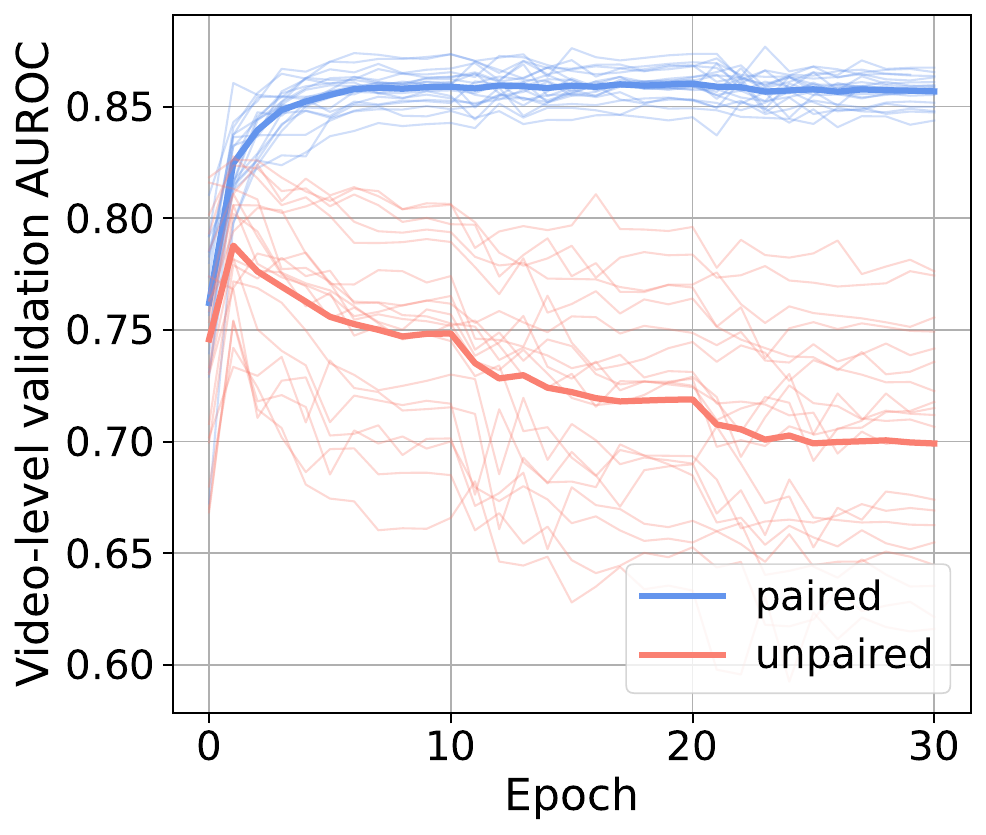}
    \end{subfigure}
       \caption{
        Video-level AUROC for (a) Training and (b) Validation, averaged over 20 randomly sampled \textcolor{paired}{paired} and 20 \textcolor{unpaired}{unpaired} datasets from FF++. The image encoder is \PEcoreL.
    }
    \label{fig:paired_unpaired}
\end{figure}

We plot AUROC for training and validation progress of \PEcoreL image encoder in \cref{fig:paired_unpaired}. Models trained on the unpaired datasets consistently overfit more quickly and achieve a lower validation AUROC, as indicated by the sharper increase in training and the decline in the validation AUROC. In contrast, models trained using paired datasets achieve a higher validation AUROC and have a lower tendency to overfit throughout the training process. We observe the same behavior on other pre-trained backbones and training sets; see supplementary material for more experiments.

We attribute the discrepancy in validation AUROC between training on paired and unpaired datasets to shortcut learning. When the model is presented with real and fake videos from the same source, it is constrained to focus on the low-level inconsistencies and artifacts introduced by the deepfake algorithm, learning a more robust and generalizable representation left by the forgery algorithm. This experiment highlights the critical importance of data pairing as a fundamental principle for developing deepfake detectors that can generalize to unseen data.

\subsection{Evolution of detection difficulty over the years}

We hypothesize that successfully training a deepfake detection system requires exposure not only to the most recent forgery techniques but also to older ones. Relying solely on recent datasets may result in poor generalization, as detectors might overfit to the artifacts specific to modern generation methods while ignoring the broader spectrum of manipulations seen over time.

\begin{figure*}[ht]
    \centering
    \includegraphics[width=1\linewidth]{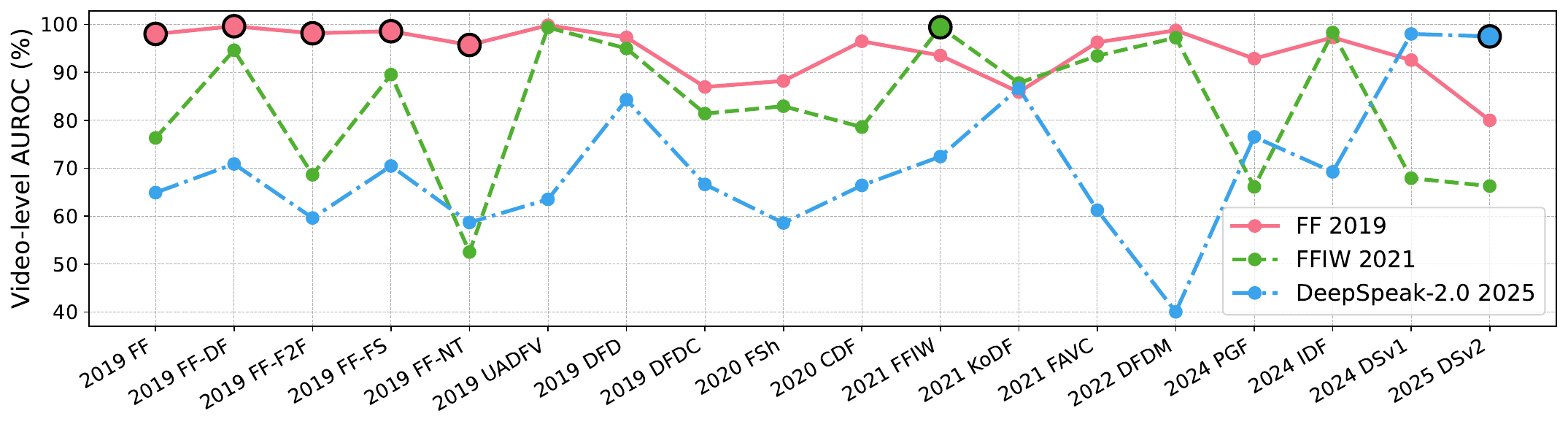}
    \caption{
        \label{fig:test_auroc_per_dataset}
        Evolution of detection difficulty over time. Each video-level AUROC is computed on the test set of the corresponding benchmark. Each curve represents a model trained on a single dataset. Highlighted circles indicate the model’s in-dataset performance.
    }
\end{figure*}

To test this, we trained three models on datasets introduced in different years: FF++ (2019), FFIW (2021), and DSv2 (2025). These models were then evaluated on a diverse set of benchmarks spanning from 2019 to 2025, with test AUROC plotted against dataset release year in~\cref{fig:test_auroc_per_dataset}.

Although the model trained on the DSv2 dataset performs well on its own test set, it generalizes poorly to older benchmarks such as FF++, DFDM, and CDFv2. In contrast, the model trained on the older but diverse FF++ dataset achieves competitive or even superior performance across a wider range of test datasets, including many recent ones.

These results indicate that training on only recent deepfakes does not guarantee a strong generalization. Instead, models trained on older datasets with diverse manipulation techniques often perform more robustly across time. This suggests that newer deepfakes do not entirely subsume or replace the challenges posed by earlier generation techniques.

To build a generalizable deepfake detector, it is crucial to curate training datasets that span both old and new forgery techniques. Focusing exclusively on recent data risks overfitting to current trends and missing broader forgery patterns.

\subsection{Robustness to image degradations}

\begin{figure*}[t]
    \centering
     \begin{subfigure}[b]{0.33\textwidth}
        \centering
        \includegraphics[width=\linewidth]{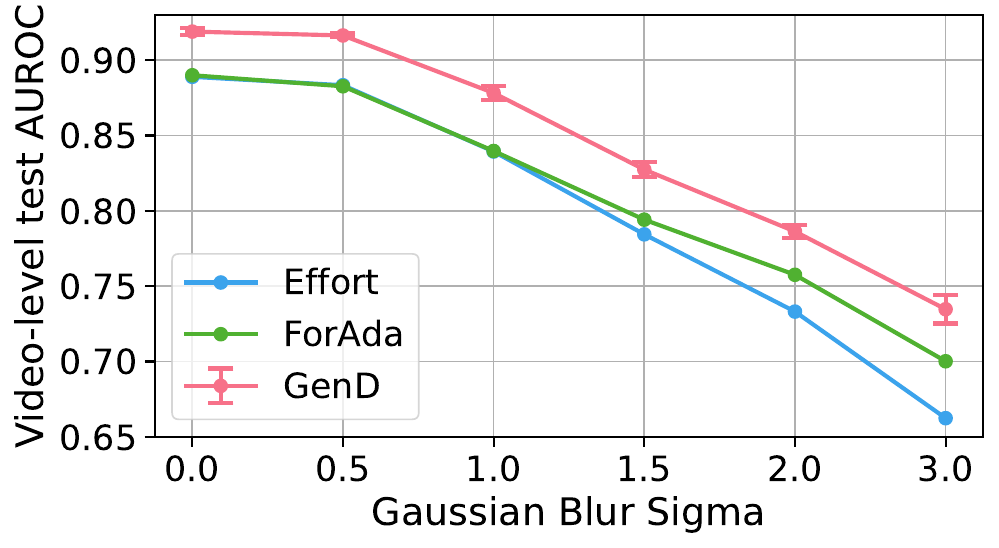}
    \end{subfigure}
    \hfill
    \begin{subfigure}[b]{0.33\textwidth}
        \centering
        \includegraphics[width=\linewidth]{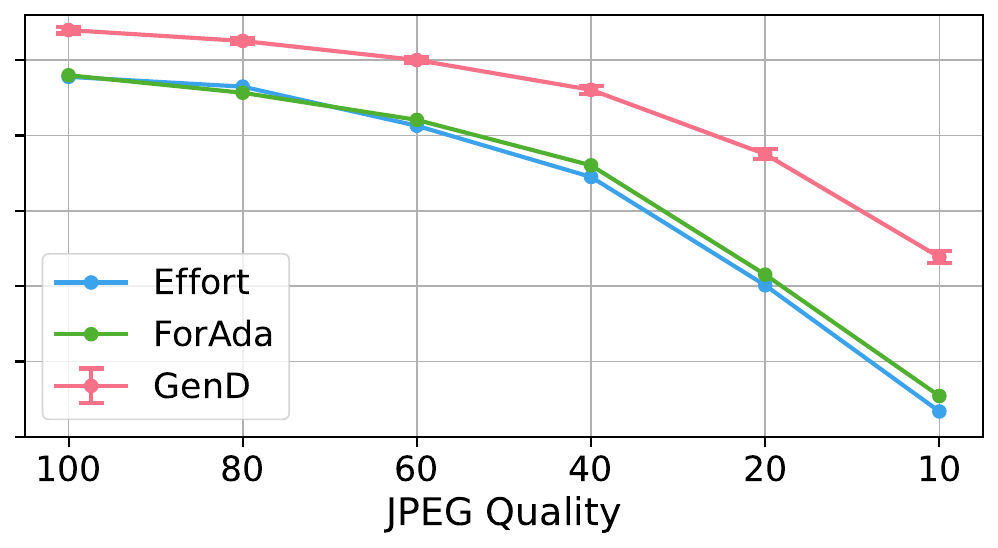}
    \end{subfigure}
    \hfill
    \begin{subfigure}[b]{0.33\textwidth}
        \centering
        \includegraphics[width=\linewidth]{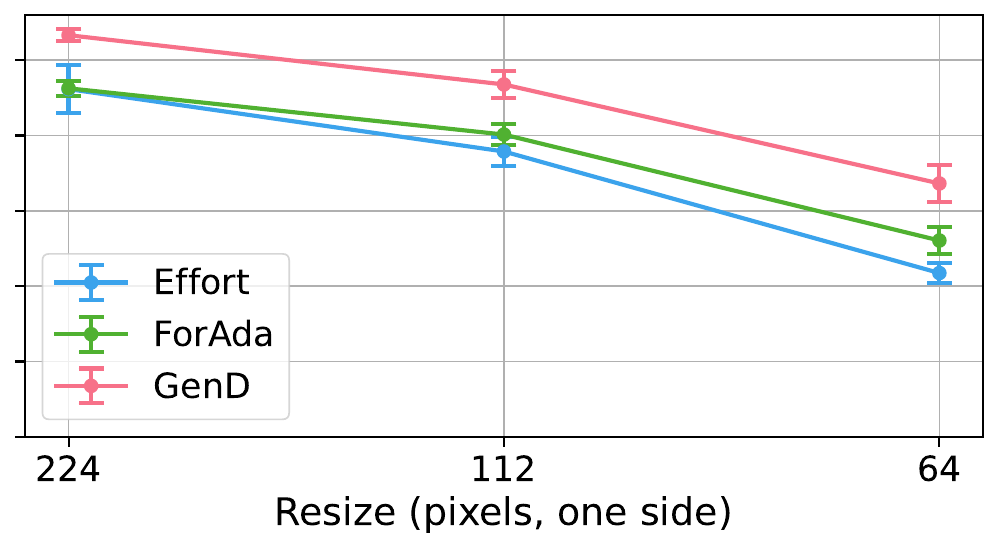}
    \end{subfigure}
   \caption{
        Robustness to image degradations for \method (\PEcoreL), ForAda~\cite{ForensicsAdapter}, and Effort~\cite{Effort}. Video-level AUROC (\%) is calculated across all 14 test datasets. Error bars for \method are computed from models trained with 5 different training seeds. In the resize, we also average every method across 6 interpolation strategies: nearest, lanczos, bilinear, bicubic, box, hamming.
    }
    \label{fig:robusteness_to_augs}
\end{figure*}

We evaluated the robustness of the \method to commonly used image degradations, which can be used to remove forgery patterns and decrease the performance of deepfake detection. For statistical significance, we average the results of \method trained using 5 different training seeds.  The results are shown in \cref{fig:robusteness_to_augs}. Visual examples of images processed over various levels of degradations can be found in the supplementary.

We selected three methods, such as Gaussian blur, JPEG compression, and image resizing. For Gaussian blur, we varied the sigma $\sigma$ and selected the proportional kernel size $k = 2 \cdot \left\lceil 3\sigma \right\rceil + 1$. For resizing, we took the facial image and resized it to $224^2$, $114^2$, or $64^2$ pixels by using 6 different interpolation techniques: nearest, lanczos, bilinear, bicubic, box, hamming. After that, each downsized image was upsampled to the model's input resolution by every method's preprocessing technique. Results are averaged over the interpolation methods.

\section{Limitations and Future Work}
Although the proposed method, \method, demonstrates a strong generalization, it is important to acknowledge its limitations, which also present avenues for future research.

\myparagraph{Robustness to perturbations and adversarial attacks.}
One limitation is that the model's performance against targeted adversarial attacks~\cite{adversarial} specifically designed to evade detection has not been evaluated. 

\myparagraph{Limited temporal modeling.}
We tried to model the temporal signal using a simple single-layer self-attention block between classification tokens from different frames in a video. Nevertheless, our experiments did not show any noticeable improvements compared to the proposed method, which operates at the frame level and aggregates predictions via simple averaging. Although this allows for architectural simplicity, it ignores temporal dynamics that could offer additional forgery cues, such as inconsistencies in lip synchronization or frame-level motion artifacts. 

\myparagraph{Dependence on facial region.}
Our preprocessing pipeline assumes access to a detectable and alignable face. Videos with occlusions, extreme poses, or low resolution may lead to failures in face detection or misaligned crops, which can degrade performance.

\myparagraph{Limited demographic diversity in training data.}
Training data lacks sufficient diversity in terms of identity, ethnicity, age, and gender. This can result in biased detection performance, where the model may generalize poorly to underrepresented demographic groups. Ensuring demographic balance and evaluating fairness across subgroups are important next steps toward responsible deployment. See the supplementary for typical failure cases. 

\myparagraph{Incremental learning.}
The current framework does not address the challenge of incremental learning. As new forgery methods emerge, the model would require complete retraining. Developing a framework that can adapt to new manipulation techniques without catastrophically forgetting previously learned ones is a critical direction for real-world applicability.

\section{Conclusions}

We addressed the persistent challenge of generalization in deepfake detection. Although recent approaches explored increasingly complex architectural and parameter-space modifications to adapt large foundation models, we show that generalization is achieved by tuning existing parameters of the image encoder (CLIP, PE, DINO). We demonstrated that such encoder is transformed into a state-of-the-art deepfake detection model using three additional components: LN-tuning, L2 normalization, and a linear classifier. State-of-the-art results are achieved within a few hours of training on a single A100 GPU, making the proposed approach computationally efficient and reproducible. 

Our extensive evaluation, conducted on one of the most comprehensive collections of deepfake benchmarks to date, confirms the efficacy of this approach, consistently matching or outperforming more complex state-of-the-art methods. Beyond performance, our work provides several insights for the community. We experimentally demonstrated that: 1) the difficulty of academic datasets did not substantially increase over time, and training on old data still gives strong generalization capabilities, compared to, for example, training on only recent datasets, and 2) training with paired real-fake data from the same source video is critical for mitigating shortcut learning and achieving better generalization. These findings indicate that data pairing and training set diversity are key factors in the development of generalizable deepfake detectors.

\myparagraph{Acknowledgments.}
The research was supported by the National Recovery Plan project CEDMO 2.0 NPO (MPO 60273/24/21300/21000), the EC Digital Europe Programme project CEDMO 2.0
no.\ 101158609, the CTU Student Grant SGS23/173/OHK3/3T/13, and the OP VVV project CZ.02.1.01/0.0/0.0/16\_019/0000765 RCI.

{
\small
\bibliographystyle{ieeenat_fullname}
\bibliography{main}
}

\clearpage
\twocolumn[
\begin{center}
    {\LARGE \textbf{Supplementary Material}}
    \vspace{1.5em}
\end{center}
]
\renewcommand{\thefigure}{S\arabic{figure}}
\renewcommand{\thetable}{S\arabic{table}}
\renewcommand{\theequation}{S\arabic{equation}}
\renewcommand{\thesection}{S\arabic{section}}

\setcounter{section}{0}
\setcounter{table}{0}
\setcounter{equation}{0}
\setcounter{figure}{0}

\section{Quantitative Results for Paired vs.\ Unpaired Training}

This section provides detailed quantitative results to support the analysis in Section 4.4 of the main paper, which argues for the importance of training on paired datasets. Although Fig. 3 in the main paper illustrates the training dynamics, \cref{tab:paired-unpaired-cross-dataset} presents the final cross-dataset generalization performance. The results are averaged over 20 different paired and 20 unpaired training sessions. 

The experiment provides strong empirical evidence for the hypothesis. The model trained on the Paired dataset consistently outperforms the one trained on the Unpaired dataset. It achieves a mean AUROC of 90.0\%, a notable improvement by 4.7 pp. This performance gain is consistent across all benchmarks.

\begin{figure}[t]
    \centering
     \begin{subfigure}[b]{0.23\textwidth}
        \centering
        \caption{
            Training
        }
        \includegraphics[width=\linewidth]{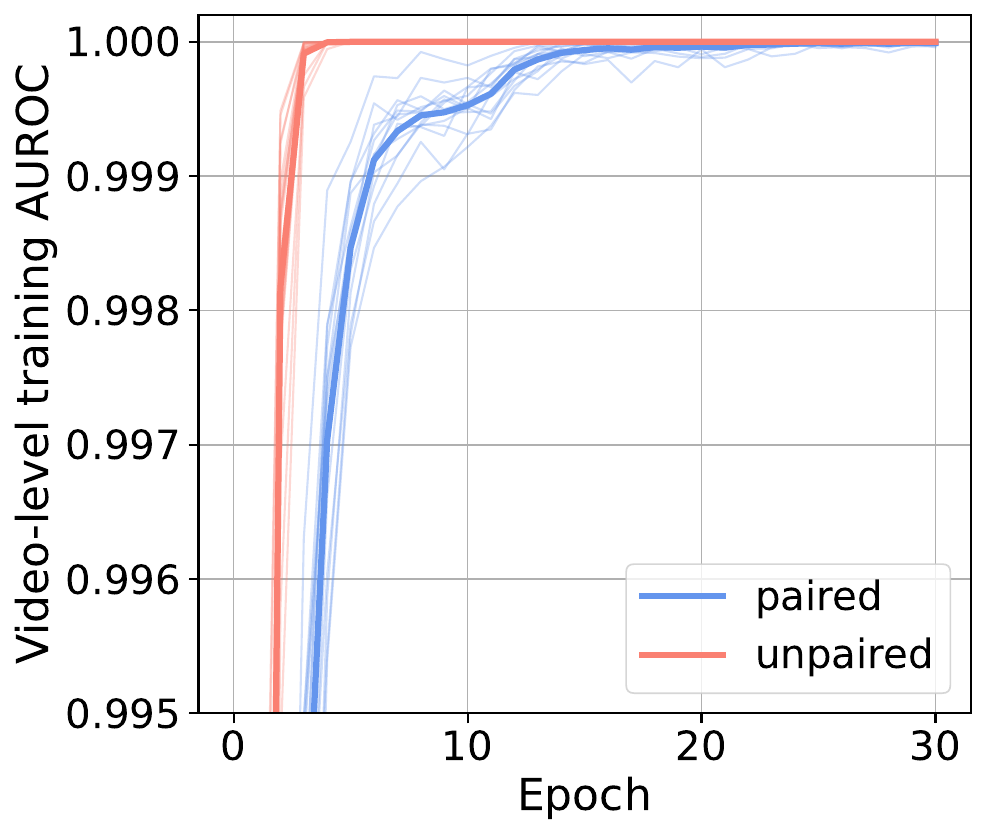}
    \end{subfigure}
    \hfill
    \begin{subfigure}[b]{0.23\textwidth}
        \centering
        \caption{
            Validation
        }
        \includegraphics[width=\linewidth]{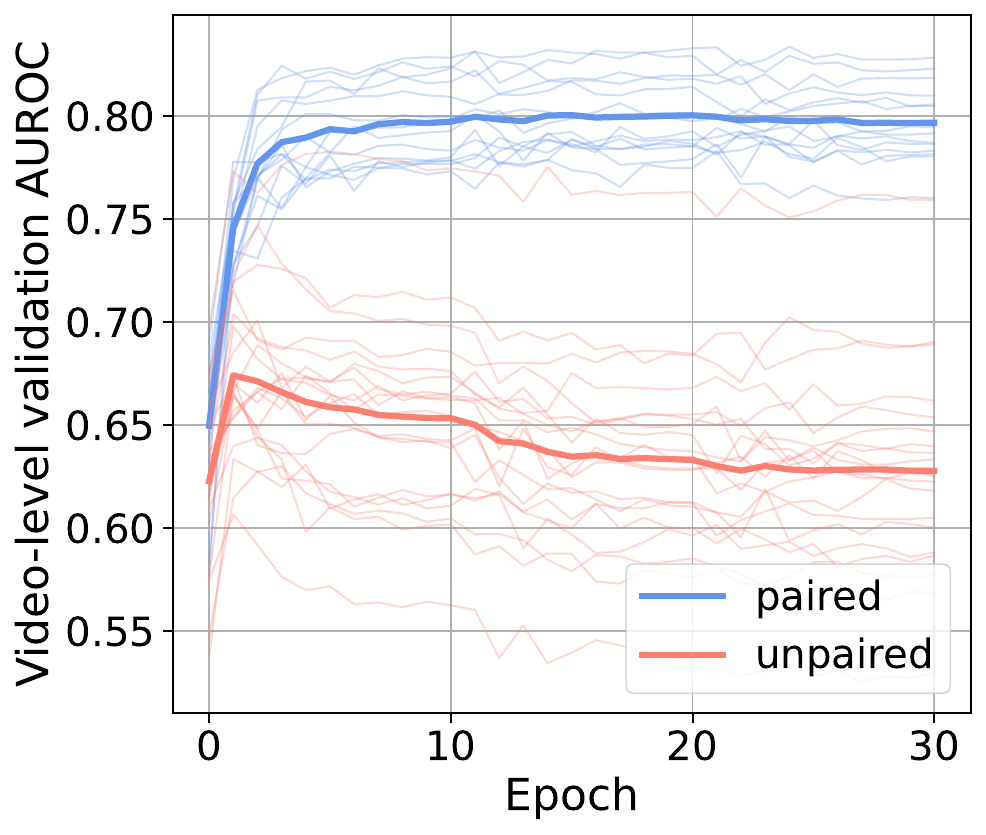}
    \end{subfigure}
    
    \caption{
    \label{fig:paired_unpaired_CLIP_FF}
     Video-level AUROC for (a) Training and (b) Validation,
averaged over 20 randomly sampled paired and unpaired datasets from the FF++ training set. The image encoder is CLIP ViT-L/14.
    }
\end{figure}

We verify that this finding generalizes to other backbones such as CLIP ViT-L/14~\cite{CLIP}, presented in \cref{fig:paired_unpaired_CLIP_FF}, where the training dynamic resembles Fig. 3 from the main paper.

\begin{figure}[t]
    \centering
     \begin{subfigure}[b]{0.23\textwidth}
        \centering
        \caption{
            Training
        }
        \includegraphics[width=\linewidth]{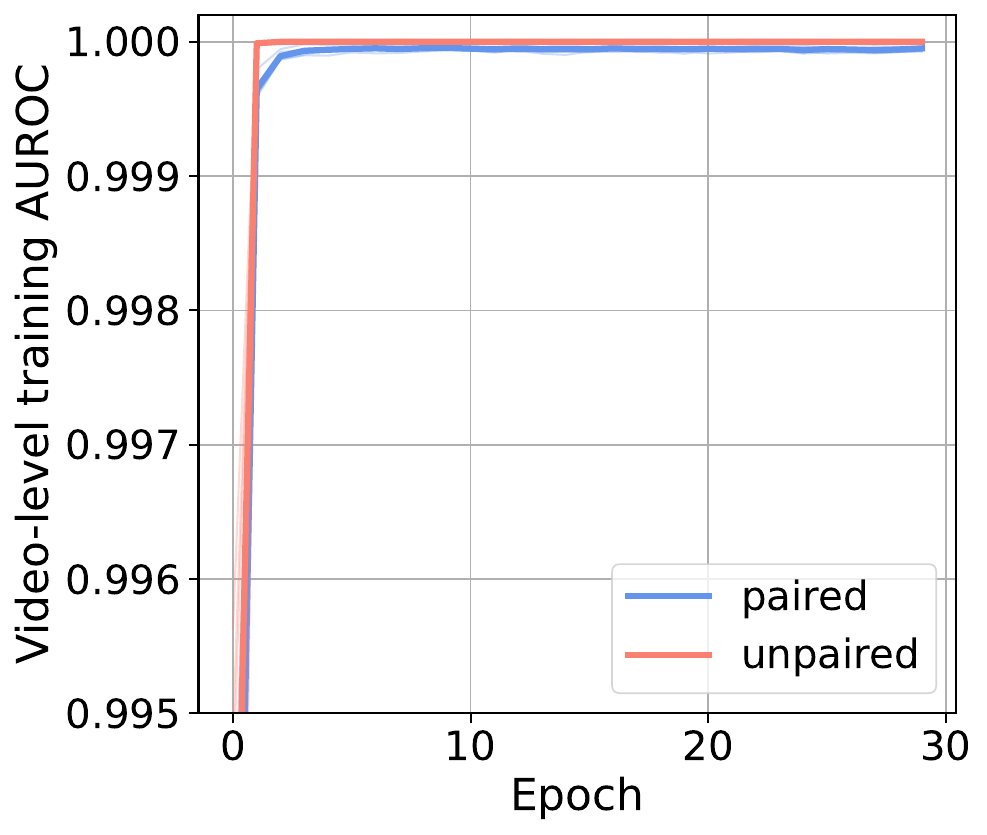}
    \end{subfigure}
    \hfill
    \begin{subfigure}[b]{0.23\textwidth}
        \centering
        \caption{
            Validation
        }
        \includegraphics[width=\linewidth]{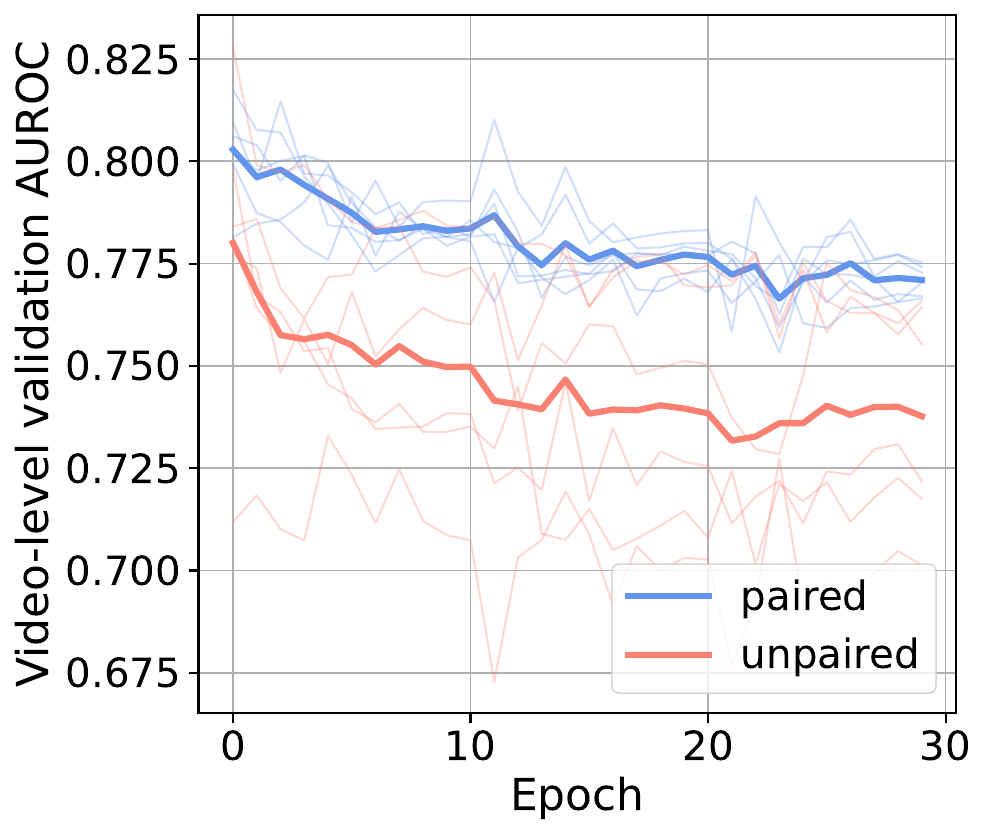}
    \end{subfigure}
    
    \caption{
    \label{fig:paired_unpaired_CLIP_FACV}
     Video-level AUROC for (a) Training and (b) Validation,
averaged over 6 randomly sampled paired and 6 unpaired datasets from the FAVC~\cite{FakeAVCeleb} dataset. The image encoder is CLIP ViT-L/14.
    }
\end{figure}

\begin{figure}[t]
    \centering
     \begin{subfigure}[b]{0.23\textwidth}
        \centering
        \caption{
            Training
        }
        \includegraphics[width=\linewidth]{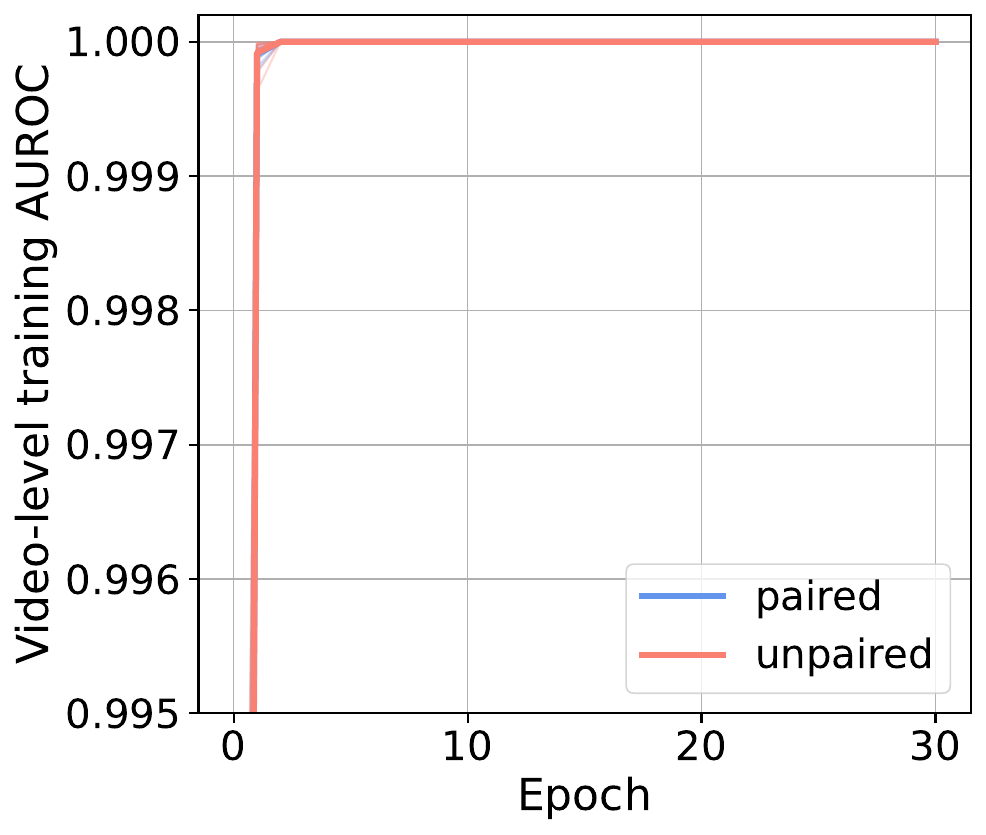}
    \end{subfigure}
    \hfill
    \begin{subfigure}[b]{0.23\textwidth}
        \centering
        \caption{
            Validation
        }
        \includegraphics[width=\linewidth]{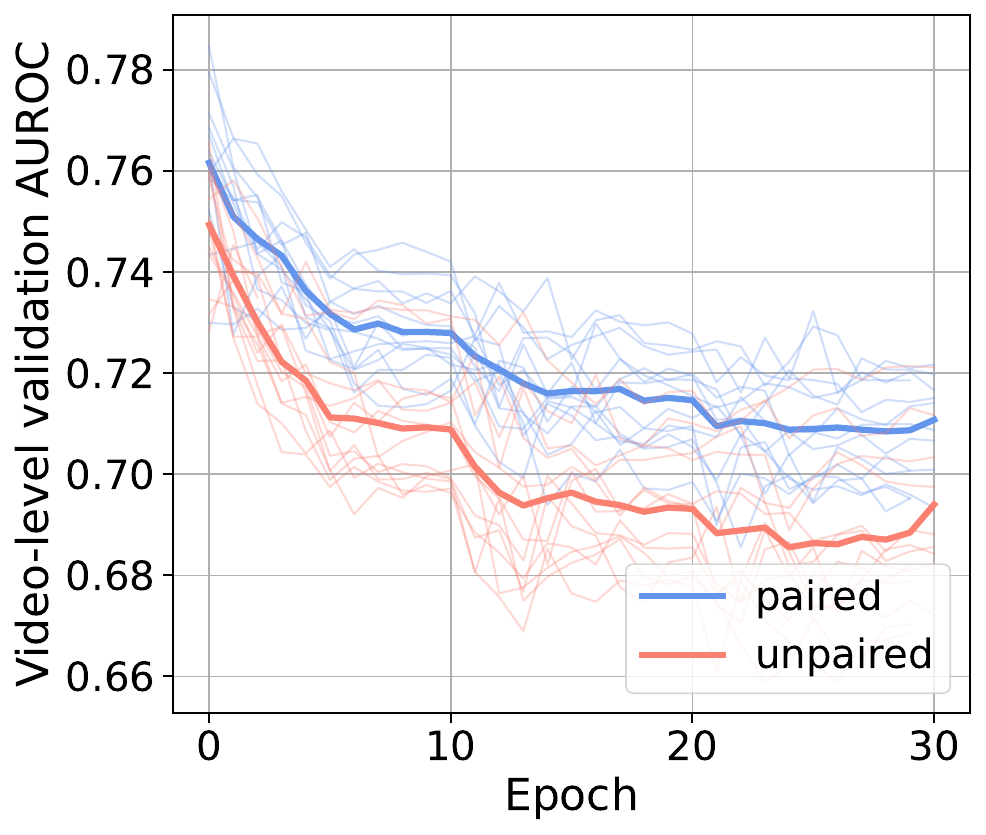}
    \end{subfigure}
    \caption{
    \label{fig:paired_unpaired_CLIP_CDFv2}
     Video-level AUROC for (a) Training and (b) Validation,
averaged over 15 randomly sampled paired and 15 unpaired datasets from the CDFv2~\cite{CDFv2} training set. The image encoder is CLIP ViT-L/14.
    }
\end{figure}

In addition, we verify that this finding generalizes to other training datasets such as CDFv2~\cite{CDFv2}, shown in \cref{fig:paired_unpaired_CLIP_CDFv2} and FAVC~\cite{FakeAVCeleb}, shown in \cref{fig:paired_unpaired_CLIP_FACV}. We observe the same training dynamics but with faster overfitting in both cases, suggesting that on our validation set, models trained on CDFv2 or FAVC training sets have lower generalization performance; see \cref{fig:paired_unpaired_CLIP_CDFv2} (b) and \cref{fig:paired_unpaired_CLIP_FACV} (b).

\begin{table*}[t]
    \centering
    \caption{
    \label{tab:paired-unpaired-cross-dataset}
     Cross-dataset video-level AUROC (\%) results for models trained on paired and unpaired datasets.
    }
    \tabcolsep=3.5pt
    \resizebox{\textwidth}{!}{
    \begin{tabular}{r|cccccccccccccc|c}
        \toprule
         \thead{Training \\ dataset}&\textbf{UADFV}& \textbf{DFD}  & \textbf{DFDC}    &\textbf{FSh}&\textbf{CDFv2} & \textbf{FFIW}  &\textbf{KoDF} &\textbf{FAVC} &\textbf{DFDM} &\textbf{PGF} &\textbf{IDF}& \textbf{DSv1.1}&\textbf{DSv2}&\textbf{CDFv3} &\textbf{Mean}\\
        \midrule
        Unpaired & 97.5$\pm$0.3 & 92.4$\pm$0.7 & 79.7$\pm$0.8 & 85.3$\pm$2.4 & 81.0$\pm$2.8 & 87.4$\pm$1.7 & 78.5$\pm$4.0 & 89.8$\pm$2.7 & 94.6$\pm$2.9 & 83.9$\pm$0.6 & 95.7$\pm$0.6 & 78.4$\pm$0.0 & 70.2$\pm$2.4 & 80.2$\pm$1.5 & 85.3 \\
        Paired & \textbf{98.1$\pm$0.3} & \textbf{95.3$\pm$0.8} & \textbf{80.4$\pm$0.2} & \textbf{86.3$\pm$1.3} & \textbf{91.5$\pm$1.3} & \textbf{90.9$\pm$1.0} & \textbf{84.8$\pm$1.6} & \textbf{95.8$\pm$0.3} & \textbf{97.8$\pm$1.0} & \textbf{91.5$\pm$2.7} & \textbf{97.0$\pm$1.2} & \textbf{87.7$\pm$1.3} & \textbf{74.6$\pm$0.2} & \textbf{88.3$\pm$2.1} & \textbf{90.0} \\
        \bottomrule
    \end{tabular}
    }
    \label{tab:paired-unpaired-results}
\end{table*}

\section{Detailed Ablation of Parameter-Efficient Fine-Tuning (PEFT) Methods}

We provide a more comprehensive evaluation of parameter-efficient fine-tuning (PEFT) strategies in \cref{tab:peft_experiments}, which were discussed in Section 4.3 of the main paper. This table evaluates and compares the cross-dataset generalization performance of Full Fine-Tuning (FFT), a baseline (where only a linear classifier is trained), and three distinct PEFT methods: BitFit~\cite{BitFit}, LoRA~\cite{hu2022lora}, and LN-tuning~\cite{LN-tuning}. Each PEFT method is evaluated in isolation as well as in combination with the other components of the proposed method: L2 normalization (+L2), uniformity and alignment losses (+UA). Every training run has the same training set and differs only in the ablated component.

As shown in the table, our findings reinforce the conclusions from the main text:

\begin{enumerate}
    \item Full Fine-Tuning (FFT) results in poor generalization, achieving the lowest mean AUROC (56.8\%) across all configurations. This supports the observation that FFT leads to rapid overfitting on the training data, as reported by, e.g.,~\cite{Effort}.

    \item All three PEFT methods substantially outperform both the FFT and the baseline, demonstrating the effectiveness of parameter-efficient adaptation for this task.

    \item Although all PEFT methods initially perform well, their synergy with the proposed components varies. We observe that LN-Tuning, together with L2 normalization and UA losses, achieves the highest mean AUROC (92.1\%); see the last row.
\end{enumerate}

This detailed comparison confirms that LN-tuning is the most effective PEFT for the problem. Striking the right balance between expressiveness and regularization, enabling synergistic improvements when combined with L2 normalization, metric learning losses, achieves the state-of-the-art generalization.

\section{Detailed performance metrics: AUROC, AP and EER, TPR@FPR=1\%,5\%}

We include additional performance metrics such as average precision (AP) in \cref{tab:results_AP}, equal error rate (EER) in \cref{tab:results_EER}, TPR@FPR=5\% in \cref{tab:results_TPR_FPR_0_05}, and TPR@FPR=1\% in \cref{tab:results_TPR_FPR_0_01} for \method, ForAda~\cite{ForensicsAdapter} and Effort~\cite{Effort} computed on all 14 cross-dataset test benchmarks presented in the paper. For the \method, we show the mean and standard deviation calculated in five different training seeds. Extending the Tab. 4 of the main paper, we include standard deviations to \cref{tab:results_AUROC}.

\section{Ablation study of L2 normalizaion and uniformity-alignment without LN-tuning}

LN-tuning~\cite{LN-tuning, LN-tuning-1, LN-tuning-2} has the most noticeable influence on the performance, allowing for the reshaping of the feature space of the classification token. This makes it more suitable for solving the deepfake detection problem using a linear classification layer. Considering that uniformity and alignment (UA) loss operates in the feature space of L2-normalized classification token features, disabling LN tuning blocks all gradient propagation from the classifier backward, rendering UA useless. This explains why rows 2 and 3 of the \cref{tab:ablation-full} lead to the same results. However, only by making the feature space hyperspherical, performance can be improved substantially for some datasets, such as IDF (15.3 pp), DFDM (6.9 pp), leading to an improvement in 8 of 14 datasets.

\section{Visual examples for image degradations}

\begin{figure*}[t]
    \centering
    
    \includegraphics[width=\linewidth]{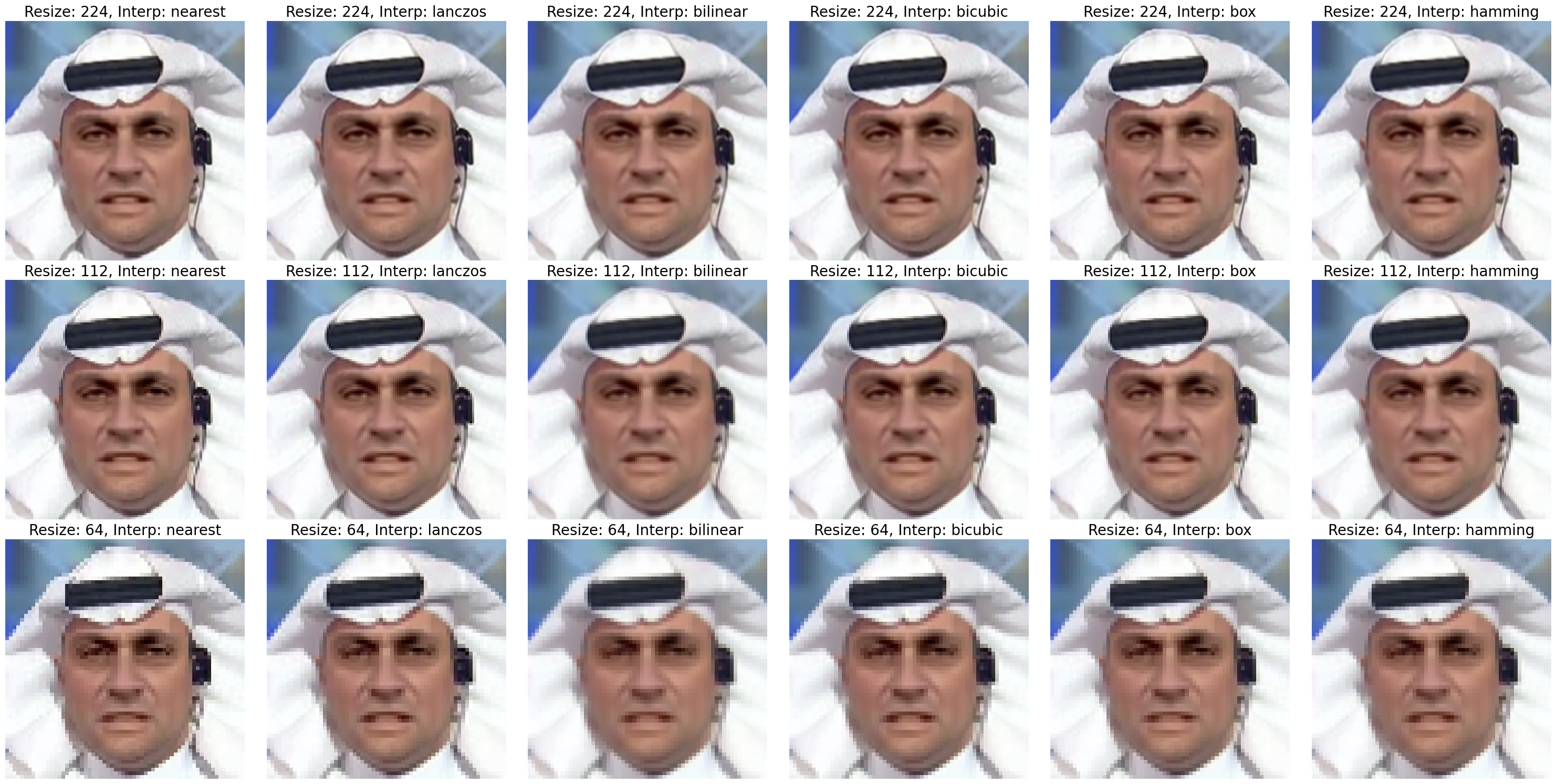}
    
    \caption{\label{fig:resize_faces}
    Visual examples for various resizing levels and interpolations used in robustness to image degradation experiments.
    }
\end{figure*}

\begin{figure*}[t]
    \centering
    
    \includegraphics[width=\linewidth]{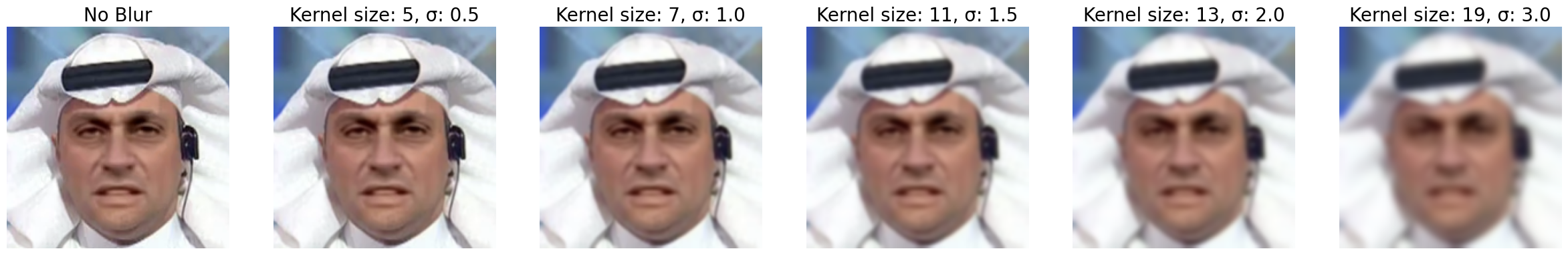}
    
    \caption{\label{fig:blur_faces}
        Visual examples for different kernel sizes and sigmas for Gaussian blurring in robustness to image degradation experiments.
    }
\end{figure*}

\begin{figure*}[t]
    \centering
    
    \includegraphics[width=\linewidth]{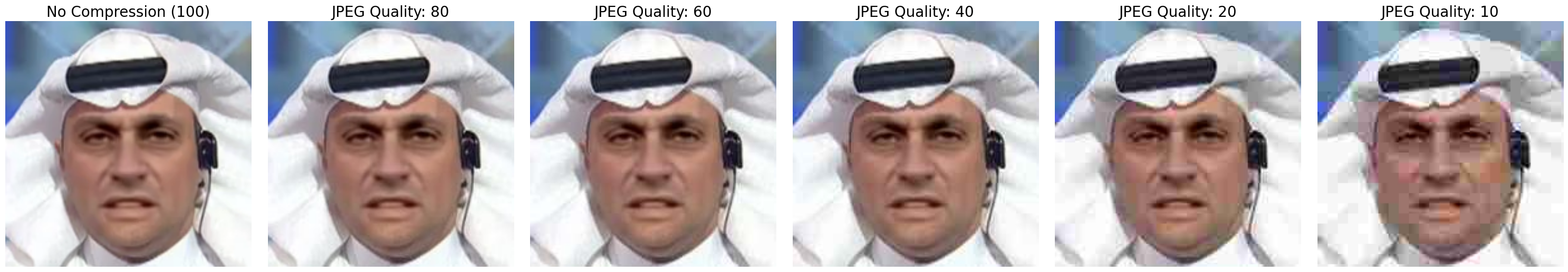}
    
    \caption{\label{fig:jpeg_faces}
    Visual examples for various JPEG compression levels in robustness to image degradation experiments.
    }
\end{figure*}

We include visualizations of different levels of image degradation, corresponding to Fig. 5 of the main paper: Resizing (\cref{fig:resize_faces}), Gaussian blurring (\cref{fig:blur_faces}), and JPEG compression (\cref{fig:jpeg_faces}).

\section{Visual examples for common failure cases}

\begin{figure*}[t]
    \centering
     \begin{subfigure}[b]{0.180\textwidth}
        \centering
        \caption{
            DFDC, F, p(F)=0.0026
        }
        \includegraphics[width=\linewidth]{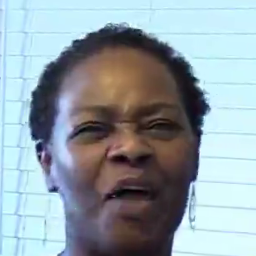}
    \end{subfigure}
    \hfill
    \begin{subfigure}[b]{0.180\textwidth}
        \centering
        \caption{
            DFDC, F, p(F)=0.0154
        }
        \includegraphics[width=\linewidth]{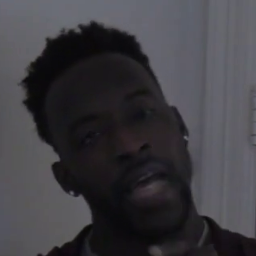}
    \end{subfigure}
    \hfill
    \begin{subfigure}[b]{0.180\textwidth}
        \centering
        \caption{
            FFIW, F, p(F)=0.0447
        }
        \includegraphics[width=\linewidth]{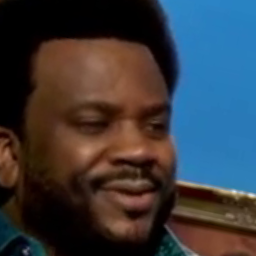}
    \end{subfigure}
    \hfill
    \begin{subfigure}[b]{0.180\textwidth}
        \centering
        \caption{
            DFDC, F, p(F)=0.9841
        }
        \includegraphics[width=\linewidth]{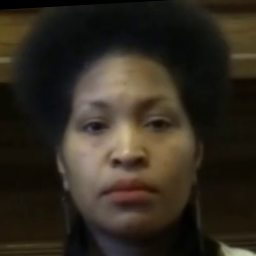}
    \end{subfigure}
    \hfill
    \begin{subfigure}[b]{0.180\textwidth}
        \centering
        \caption{
            DFDC, F, p(F)=0.8872
        }
        \includegraphics[width=\linewidth]{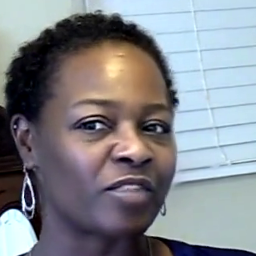}
    \end{subfigure}
    \hfill
    \begin{subfigure}[b]{0.180\textwidth}
        \centering
        \caption{
            DSv2, F, p(F)=0.0042
        }
        \includegraphics[width=\linewidth]{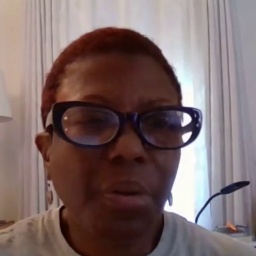}
    \end{subfigure}
    \hfill
    \begin{subfigure}[b]{0.180\textwidth}
        \centering
        \caption{
            DSv2, F, p(F)=0.0170
        }
        \includegraphics[width=\linewidth]{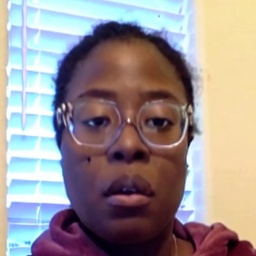}
    \end{subfigure}
    \hfill
    \begin{subfigure}[b]{0.180\textwidth}
        \centering
        \caption{
            DSv2, F, p(F)=0.2160
        }
        \includegraphics[width=\linewidth]{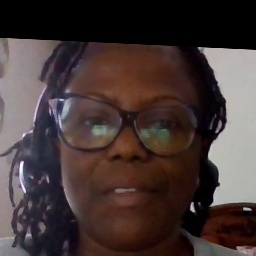}
    \end{subfigure}
    \hfill
    \begin{subfigure}[b]{0.180\textwidth}
        \centering
        \caption{
            FFIW, F, p(F)=0.0967
        }
        \includegraphics[width=\linewidth]{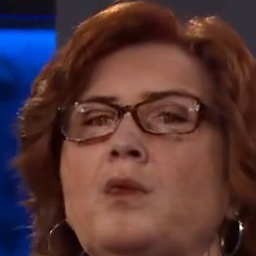}
    \end{subfigure}
    \hfill
    \begin{subfigure}[b]{0.180\textwidth}
        \centering
        \caption{
            DSv2, F, p(F)=0.0792
        }
        \includegraphics[width=\linewidth]{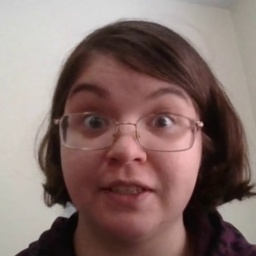}
    \end{subfigure}
    \hfill
    \begin{subfigure}[b]{0.180\textwidth}
        \centering
        \caption{
            DSv1.1, R, p(F)=0.9792
        }
        \includegraphics[width=\linewidth]{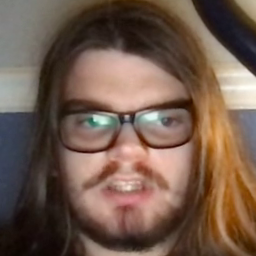}
    \end{subfigure}
    \hfill
    \begin{subfigure}[b]{0.180\textwidth}
        \centering
        \caption{
            DSv2, R, p(F)=0.9526
        }
        \includegraphics[width=\linewidth]{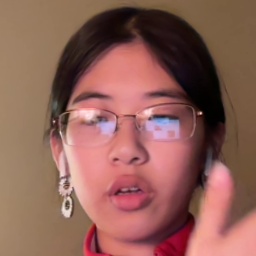}
    \end{subfigure}
    \hfill
    \begin{subfigure}[b]{0.180\textwidth}
        \centering
        \caption{
            DSv2, F, p(F)=0.0029
        }
        \includegraphics[width=\linewidth]{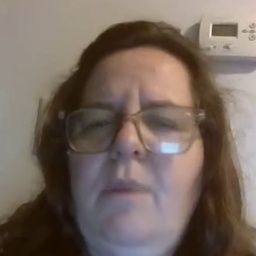}
    \end{subfigure}
    \hfill
    \begin{subfigure}[b]{0.180\textwidth}
        \centering
        \caption{
            KoDF, F, p(F)=0.0091
        }
        \includegraphics[width=\linewidth]{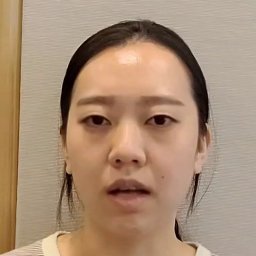}
    \end{subfigure}
    \hfill
    \begin{subfigure}[b]{0.180\textwidth}
        \centering
        \caption{
            KoDF, R, p(F)=0.9838
        }
        \includegraphics[width=\linewidth]{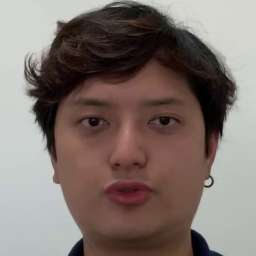}
    \end{subfigure}
    \hfill
    \begin{subfigure}[b]{0.180\textwidth}
        \centering
        \caption{
            FFIW, R, p(F)=0.9315
        }
        \includegraphics[width=\linewidth]{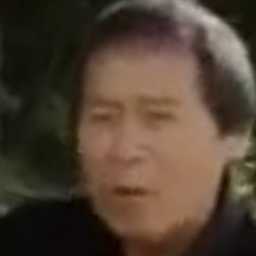}
    \end{subfigure}
    \hfill
    \begin{subfigure}[b]{0.180\textwidth}
        \centering
        \caption{
            DSv2, R, p(F)=0.9454
        }
        \includegraphics[width=\linewidth]{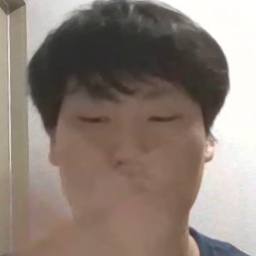}
    \end{subfigure}
    \hfill
    \begin{subfigure}[b]{0.180\textwidth}
        \centering
        \caption{
            DFDC, R, p(F)=0.9917
        }
        \includegraphics[width=\linewidth]{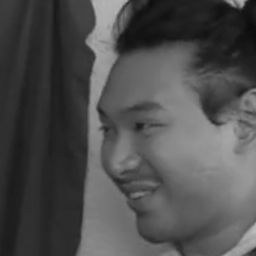}
    \end{subfigure}
    \hfill
    \begin{subfigure}[b]{0.180\textwidth}
        \centering
        \caption{
            FFIW, R, p(F)=0.9906
        }
        \includegraphics[width=\linewidth]{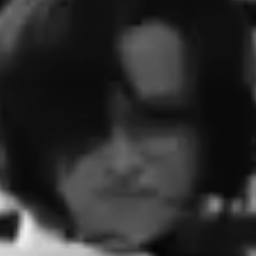}
    \end{subfigure}
    \hfill
    \begin{subfigure}[b]{0.180\textwidth}
        \centering
        \caption{
            DFDC, R, p(F)=0.9914
        }
        \includegraphics[width=\linewidth]{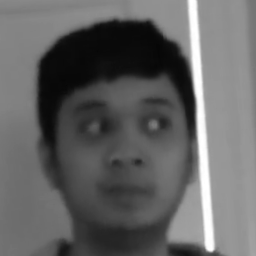}
    \end{subfigure}
    \hfill
    \begin{subfigure}[b]{0.180\textwidth}
        \centering
        \caption{
            DFDC, F, p(F)=0.1128
        }
        \includegraphics[width=\linewidth]{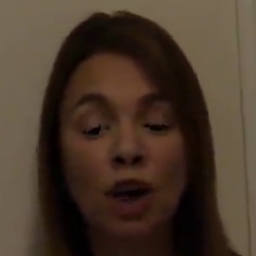}
    \end{subfigure}
    \hfill
    \begin{subfigure}[b]{0.180\textwidth}
        \centering
        \caption{
            DFDC, R, p(F)=0.9797
        }
        \includegraphics[width=\linewidth]{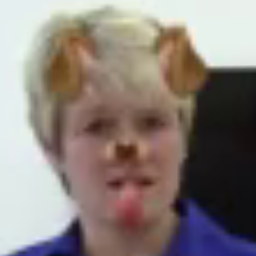}
    \end{subfigure}
    \hfill
    \begin{subfigure}[b]{0.180\textwidth}
        \centering
        \caption{
            DFDC, R, p(F)=0.9906
        }
        \includegraphics[width=\linewidth]{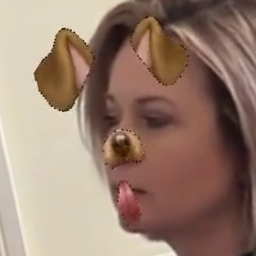}
    \end{subfigure}
    \hfill
    \begin{subfigure}[b]{0.180\textwidth}
        \centering
        \caption{
            DFDC, R, p(F)=0.8800
        }
        \includegraphics[width=\linewidth]{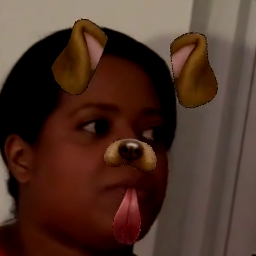}
    \end{subfigure}
    \hfill
    \begin{subfigure}[b]{0.180\textwidth}
        \centering
        \caption{
            DFDC, R, p(F)=0.9955
        }
        \includegraphics[width=\linewidth]{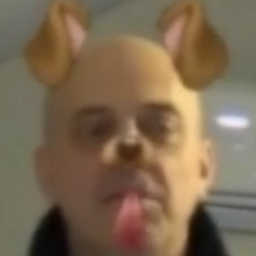}
    \end{subfigure}
    
    \caption{
    \label{fig:failure-cases}
     Examples of common failure cases. The first word denotes the dataset. The second word is the ground truth class, R for real and F for fake. p(F) means the predicted probability of a frame being of a fake class.
    }
\end{figure*}

We visually present common failure cases in \cref{fig:failure-cases}. We ordered approximately 60K testing videos from the most to the least misclassified video according to the softmax score of the output. We manually investigated the top 300 videos with the highest error. We observe a few distinct modes of failures. Photos a-h show that the network misclassified black people. Photos f-m suggest that eyeglasses may cause misclassifications. Photos n-r again signify the ethnic bias for asian people. Photos p-u can be failures due to low quality, image intensity, contrast, or monochrome processing. An interesting case is v-y, showing that the ground-truth class represents a real category, but the photos have been visibly altered through the addition of cartoon-like effects.

\begin{table*}[t]
    \centering
    \caption{\label{tab:peft_experiments}
        Experimental results comparing various PEFT strategies for CLIP ViT-L/14 backbone. Each row represents a different model configuration across all test datasets. The best video-level AUROC results are shown in bold. The last row is the proposed \method.}
    \resizebox{\textwidth}{!}{%
    \begin{tabular}{l|ccccccccccccc|c}
        \toprule
         \textbf{Method} & \thead{2019 \\ \textbf{UADFV}} & \thead{2019 \\ \textbf{DFD}} & \thead{2019 \\ \textbf{DFDC}} & \thead{2020 \\ \textbf{FSh}} & \thead{2020 \\ \textbf{CDFv2}} & \thead{2021 \\ \textbf{FFIW}} & \thead{2021 \\ \textbf{KoDF}} & \thead{2021 \\ \textbf{FAVC}} & \thead{2022 \\ \textbf{DFDM}} & \thead{2024 \\ \textbf{PGF}} & \thead{2024 \\ \textbf{IDF}} & \thead{2024 \\ \textbf{DSv1.1}} & \thead{2025 \\ \textbf{DSv2}}  & \textbf{Mean} \\
        \midrule
FFT & 62.4 & 49.5 & 56.0 & 58.8 & 65.5 & 53.4 & 52.8 & 60.8 & 59.7 & 57.5 & 68.3 & 56.7 & 50.8 & 57.9 \\
Baseline & 95.8 & 86.7 & 74.0 & 77.9 & 78.8 & 84.6 & 82.2 & 80.8 & 80.7 & 61.1 & 68.2 & 61.6 & 56.6 & 76.1 \\
BitFit & 99.6 & 96.9 & 84.6 & 86.6 & 95.3 & 91.2 & 88.9 & 95.4 & 99.6 & 86.5 & 94.2 & 86.6 & 76.0 & 90.9 \\
BitFit+L2 & 99.4 & 96.1 & 82.6 & 85.3 & 92.7 & 90.3 & 85.0 & 96.2 & 98.8 & 91.8 & \textbf{98.1} & 88.9 & 76.8 & 90.9 \\
BitFit+L2+UA & 99.8 & 96.9 & 85.9 & 85.2 & \textbf{95.4} & 89.4 & 87.6 & 96.1 & \textbf{99.9} & \textbf{92.1} & 97.0 & 90.2 & 76.3 & 91.7 \\
LoRA & 98.8 & 97.7 & 84.4 & 89.7 & 94.1 & \textbf{93.7} & 88.2 & \textbf{96.3} & 99.4 & 90.7 & 96.6 & 89.7 & 75.9 & 91.9 \\
LoRA+L2 & 98.6 & 95.5 & 81.9 & 86.2 & 89.3 & 90.8 & 87.8 & 95.5 & 99.4 & 90.6 & 95.3 & 89.4 & 75.9 & 90.5 \\
LoRA+L2+UA & 99.5 & 96.1 & 86.2 & \textbf{92.0} & 92.3 & 92.7 & \textbf{90.8} & 94.8 & 99.5 & 88.1 & 96.4 & 86.7 & 73.0 & 91.4 \\
LN & 99.3 & 96.6 & 84.0 & 86.7 & 93.1 & 91.6 & 87.4 & 94.8 & 99.1 & 87.8 & 93.3 & 89.1 & 77.6 & 90.8 \\
LN+L2 & \textbf{99.8} & 97.3 & \textbf{86.6} & 88.9 & 94.5 & 91.6 & 84.5 & 96.0 & 99.4 & 89.6 & 95.2 & 89.8 & \textbf{79.8} & 91.8 \\
LN+L2+UA & 99.6 & \textbf{97.8} & 86.5 & 87.3 & 94.3 & 90.7 & 87.0 & 96.1 & 99.6 & 90.7 & 98.0 & \textbf{90.6} & 78.6 & \textbf{92.1} \\
        \bottomrule
    \end{tabular}%
    }
\end{table*}

\clearpage

\begin{table*}[t]
    \centering
    \caption{\label{tab:results_AUROC}
    Cross-dataset video-level AUROC (\%) for reproduced methods. The highest score in each column is in bold. Results for \method are the averages over five training seeds.}

    \resizebox{\textwidth}{!}{%
    \begin{tabular}{l|cccccccccccccc|c}
        \toprule
         \textbf{Method} & \thead{2019 \\ \textbf{UADFV}} & \thead{2019 \\ \textbf{DFD}} & \thead{2019 \\ \textbf{DFDC}} & \thead{2020 \\ \textbf{FSh}} & \thead{2020 \\ \textbf{CDFv2}} & \thead{2021 \\ \textbf{FFIW}} & \thead{2021 \\ \textbf{KoDF}} & \thead{2021 \\ \textbf{FAVC}} & \thead{2022 \\ \textbf{DFDM}} & \thead{2024 \\ \textbf{PGF}} & \thead{2024 \\ \textbf{IDF}} & \thead{2024 \\ \textbf{DSv1.1}} & \thead{2025 \\ \textbf{DSv2}}  &\thead{2025  \\ \textbf{CDFv3}}& \textbf{Mean} \\
        \midrule
ForAda~\cite{ForensicsAdapter} & \textbf{99.4} & \textbf{97.2} & \textbf{87.3} & 82.0 & \textbf{95.7} & 90.6 & 88.2 & 93.1 & 97.1 & 86.6 & 90.8 & 81.8 & 72.8 & 75.6 & 88.4 \\
Effort~\cite{Effort} & 97.4 & 95.2 & 85.4 & \textbf{91.2} & 93.2 & 92.5 & 88.1 & 92.4 & 98.2 & 84.9 & 96.0 & 82.1 & 64.4 & 78.7 & 88.5 \\
GenD (CLIP) & 99.2$\pm$0.1 & 96.4$\pm$0.5 & 86.4$\pm$0.4 & 86.6$\pm$0.5 & 94.6$\pm$0.9 & 91.5$\pm$1.4 & 84.9$\pm$0.6 & 96.0$\pm$0.8 & 99.6$\pm$0.1 & 89.6$\pm$0.6 & 97.8$\pm$0.5 & \textbf{90.1$\pm$0.8} & 77.7$\pm$0.7 & 85.9$\pm$0.7 & 91.2 \\
GenD (PE) & 97.7$\pm$0.2 & 96.8$\pm$0.5 & 82.2$\pm$0.5 & 87.6$\pm$1.1 & 95.0$\pm$0.8 & \textbf{93.7$\pm$0.5} & 85.1$\pm$1.2 & 97.3$\pm$0.6 & 98.3$\pm$0.5 & 92.3$\pm$0.5 & 97.9$\pm$0.5 & 87.8$\pm$1.6 & 78.6$\pm$1.6 & \textbf{89.5$\pm$0.6} & 91.4 \\
GenD (DINO) & 98.6$\pm$0.1 & 96.2$\pm$0.4 & 85.6$\pm$0.5 & 88.8$\pm$1.3 & 92.5$\pm$0.9 & 92.9$\pm$1.2 & \textbf{89.7$\pm$0.7} & \textbf{98.4$\pm$0.5} & \textbf{99.8$\pm$0.1} & \textbf{92.4$\pm$0.4} & \textbf{98.2$\pm$0.5} & 86.9$\pm$0.8 & \textbf{79.4$\pm$0.6} & 83.5$\pm$1.0 & \textbf{91.6} \\
        \bottomrule
    \end{tabular}%
    }
\end{table*}

\begin{table*}[t]
    \centering
    \caption{\label{tab:results_EER}
    Cross-dataset video-level EER (\%) for reproduced methods. The lowest score in each column is in bold. Results for \method are the averages over five training seeds.}

    \resizebox{\textwidth}{!}{%
    \begin{tabular}{l|cccccccccccccc|c}
        \toprule
         \textbf{Method} & \thead{2019 \\ \textbf{UADFV}} & \thead{2019 \\ \textbf{DFD}} & \thead{2019 \\ \textbf{DFDC}} & \thead{2020 \\ \textbf{FSh}} & \thead{2020 \\ \textbf{CDFv2}} & \thead{2021 \\ \textbf{FFIW}} & \thead{2021 \\ \textbf{KoDF}} & \thead{2021 \\ \textbf{FAVC}} & \thead{2022 \\ \textbf{DFDM}} & \thead{2024 \\ \textbf{PGF}} & \thead{2024 \\ \textbf{IDF}} & \thead{2024 \\ \textbf{DSv1.1}} & \thead{2025 \\ \textbf{DSv2}}  &\thead{2025  \\ \textbf{CDFv3}}& \textbf{Mean} \\
        \midrule
ForAda~\cite{ForensicsAdapter} & 4.1 & \textbf{9.0} & \textbf{20.7} & 24.3 & \textbf{11.2} & 17.1 & 20.0 & 15.2 & 8.5 & 22.0 & 18.4 & 25.6 & 30.5 & 32.0 & 18.5 \\
Effort~\cite{Effort} & 6.1 & 11.3 & 22.9 & \textbf{16.4} & 14.7 & 16.1 & 18.4 & 16.2 & 7.1 & 23.2 & 11.0 & 24.6 & 40.0 & 29.8 & 18.4 \\
GenD (CLIP) & \textbf{3.7$\pm$1.7} & 10.4$\pm$0.8 & 21.8$\pm$0.3 & 21.7$\pm$0.8 & 13.2$\pm$1.4 & 16.8$\pm$1.8 & 22.7$\pm$0.8 & 11.2$\pm$1.0 & 2.7$\pm$0.6 & 17.9$\pm$1.0 & 7.7$\pm$1.2 & \textbf{16.1$\pm$0.8} & 28.7$\pm$0.8 & 22.1$\pm$0.7 & 15.5 \\
GenD (PE) & 6.5$\pm$0.9 & 9.3$\pm$0.8 & 24.8$\pm$0.7 & 20.7$\pm$1.0 & 12.0$\pm$1.3 & \textbf{13.1$\pm$0.7} & 22.6$\pm$1.2 & 9.2$\pm$1.4 & 7.2$\pm$1.2 & \textbf{14.5$\pm$0.7} & 7.9$\pm$1.0 & 20.5$\pm$2.8 & 29.8$\pm$1.3 & \textbf{17.4$\pm$0.7} & 15.4 \\
GenD (DINO) & 4.9$\pm$1.1 & 10.7$\pm$0.6 & 22.6$\pm$0.5 & 19.1$\pm$1.1 & 14.4$\pm$1.2 & 14.6$\pm$1.4 & \textbf{17.4$\pm$0.7} & \textbf{6.2$\pm$0.9} & \textbf{1.8$\pm$0.4} & 15.6$\pm$0.7 & \textbf{6.8$\pm$1.2} & 20.5$\pm$1.1 & \textbf{28.2$\pm$1.0} & 22.9$\pm$1.2 & \textbf{14.7} \\
        \bottomrule
    \end{tabular}%
    }
\end{table*}

\begin{table*}[t]
    \centering
    \caption{\label{tab:results_AP}
    Cross-dataset video-level AP (\%) for reproduced methods. The highest score in each column is in bold. Results for \method are the averages over five training seeds.}

    \resizebox{\textwidth}{!}{%
    \begin{tabular}{l|cccccccccccccc|c}
        \toprule
         \textbf{Method} & \thead{2019 \\ \textbf{UADFV}} & \thead{2019 \\ \textbf{DFD}} & \thead{2019 \\ \textbf{DFDC}} & \thead{2020 \\ \textbf{FSh}} & \thead{2020 \\ \textbf{CDFv2}} & \thead{2021 \\ \textbf{FFIW}} & \thead{2021 \\ \textbf{KoDF}} & \thead{2021 \\ \textbf{FAVC}} & \thead{2022 \\ \textbf{DFDM}} & \thead{2024 \\ \textbf{PGF}} & \thead{2024 \\ \textbf{IDF}} & \thead{2024 \\ \textbf{DSv1.1}} & \thead{2025 \\ \textbf{DSv2}}  &\thead{2025  \\ \textbf{CDFv3}}& \textbf{Mean} \\
        \midrule
ForAda~\cite{ForensicsAdapter} & \textbf{99.4} & \textbf{90.0} & \textbf{86.9} & 81.7 & \textbf{95.3} & 90.3 & 79.5 & 66.0 & 95.4 & 64.7 & 85.7 & 80.4 & 70.5 & 53.6 & 81.4 \\
Effort~\cite{Effort} & 97.2 & 82.0 & 84.6 & \textbf{90.4} & 91.9 & 92.6 & 80.6 & 61.7 & 97.0 & 66.3 & 93.1 & 80.4 & 63.1 & 54.2 & 81.1 \\
GenD (CLIP) & 99.3$\pm$0.1 & 88.8$\pm$1.5 & 86.1$\pm$0.5 & 86.2$\pm$0.4 & 94.3$\pm$0.9 & 91.5$\pm$1.4 & 74.6$\pm$0.5 & 76.7$\pm$3.8 & 99.5$\pm$0.2 & 72.7$\pm$0.8 & 96.2$\pm$0.7 & \textbf{89.2$\pm$0.7} & 76.0$\pm$0.8 & 58.5$\pm$0.8 & 85.0 \\
GenD (PE) & 97.8$\pm$0.2 & 89.1$\pm$1.5 & 81.5$\pm$0.4 & 87.8$\pm$0.9 & 94.7$\pm$0.8 & \textbf{93.5$\pm$0.5} & 75.4$\pm$1.7 & 81.5$\pm$1.8 & 97.5$\pm$0.7 & 79.7$\pm$1.4 & 96.3$\pm$0.7 & 87.6$\pm$1.6 & 78.8$\pm$1.6 & \textbf{59.8$\pm$0.6} & 85.8 \\
GenD (DINO) & 98.7$\pm$0.1 & 88.5$\pm$0.9 & 85.1$\pm$0.5 & 88.9$\pm$1.2 & 92.0$\pm$0.9 & 92.7$\pm$1.2 & \textbf{82.5$\pm$1.1} & \textbf{87.8$\pm$2.7} & \textbf{99.7$\pm$0.1} & \textbf{80.6$\pm$1.5} & \textbf{96.7$\pm$1.3} & 86.5$\pm$0.8 & \textbf{78.9$\pm$0.6} & 55.8$\pm$0.9 & \textbf{86.8} \\
        \bottomrule
    \end{tabular}%
    }
\end{table*}

\begin{table*}[t]
    \centering
    \caption{\label{tab:results_TPR_FPR_0_05}
    Cross-dataset video-level TPR@FPR=5\% for reproduced methods. The highest score in each column is in bold. Results for \method are the averages over five training seeds.}

    \resizebox{\textwidth}{!}{%
    \begin{tabular}{l|cccccccccccccc|c}
        \toprule
         \textbf{Method} & \thead{2019 \\ \textbf{UADFV}} & \thead{2019 \\ \textbf{DFD}} & \thead{2019 \\ \textbf{DFDC}} & \thead{2020 \\ \textbf{FSh}} & \thead{2020 \\ \textbf{CDFv2}} & \thead{2021 \\ \textbf{FFIW}} & \thead{2021 \\ \textbf{KoDF}} & \thead{2021 \\ \textbf{FAVC}} & \thead{2022 \\ \textbf{DFDM}} & \thead{2024 \\ \textbf{PGF}} & \thead{2024 \\ \textbf{IDF}} & \thead{2024 \\ \textbf{DSv1.1}} & \thead{2025 \\ \textbf{DSv2}}  &\thead{2025  \\ \textbf{CDFv3}}& \textbf{Mean} \\
        \midrule
ForAda~\cite{ForensicsAdapter} & \textbf{95.9} & 87.3 & 55.7 & 46.4 & \textbf{76.2} & 60.1 & 72.2 & 73.8 & 86.7 & 51.5 & 61.5 & 45.5 & 19.4 & 33.9 & 61.9 \\
Effort~\cite{Effort} & 93.9 & 84.5 & \textbf{57.0} & \textbf{74.3} & 73.8 & \textbf{72.3} & 71.4 & 73.8 & 91.7 & 51.3 & 81.5 & 45.8 & 16.9 & 42.0 & 66.4 \\
GenD (CLIP) & 94.9$\pm$1.4 & 80.9$\pm$1.1 & 50.6$\pm$0.8 & 53.6$\pm$4.0 & 68.8$\pm$0.8 & 62.6$\pm$0.8 & 69.0$\pm$1.0 & 81.5$\pm$1.8 & 98.6$\pm$0.4 & 49.9$\pm$1.0 & 88.5$\pm$2.2 & \textbf{59.5$\pm$1.5} & 31.5$\pm$0.6 & 55.0$\pm$0.3 & 67.5 \\
GenD (PE) & 91.4$\pm$2.2 & \textbf{87.5$\pm$1.9} & 32.7$\pm$1.7 & 51.9$\pm$5.2 & 72.1$\pm$4.7 & 67.7$\pm$2.6 & 67.4$\pm$1.6 & 86.7$\pm$2.6 & 90.2$\pm$3.2 & 60.2$\pm$2.6 & 89.2$\pm$2.3 & 41.3$\pm$4.6 & \textbf{42.7$\pm$4.4} & \textbf{64.3$\pm$2.0} & 67.5 \\
GenD (DINO) & 94.7$\pm$2.7 & 84.0$\pm$2.2 & 54.2$\pm$4.0 & 57.4$\pm$9.1 & 57.6$\pm$5.2 & 68.8$\pm$5.9 & \textbf{73.4$\pm$0.9} & \textbf{92.1$\pm$2.5} & \textbf{99.3$\pm$0.3} & \textbf{61.9$\pm$2.9} & \textbf{91.5$\pm$3.0} & 46.3$\pm$5.0 & 41.1$\pm$4.9 & 48.3$\pm$1.2 & \textbf{69.3} \\
        \bottomrule
    \end{tabular}%
    }
\end{table*}

\begin{table*}[t]
    \centering
    \caption{\label{tab:results_TPR_FPR_0_01}
    Cross-dataset video-level TPR@FPR=1\% for reproduced methods. The highest score in each column is in bold. Results for \method are the averages over five training seeds.}
    \resizebox{\textwidth}{!}{%
    \begin{tabular}{l|cccccccccccccc|c}
        \toprule
         \textbf{Method} & \thead{2019 \\ \textbf{UADFV}} & \thead{2019 \\ \textbf{DFD}} & \thead{2019 \\ \textbf{DFDC}} & \thead{2020 \\ \textbf{FSh}} & \thead{2020 \\ \textbf{CDFv2}} & \thead{2021 \\ \textbf{FFIW}} & \thead{2021 \\ \textbf{KoDF}} & \thead{2021 \\ \textbf{FAVC}} & \thead{2022 \\ \textbf{DFDM}} & \thead{2024 \\ \textbf{PGF}} & \thead{2024 \\ \textbf{IDF}} & \thead{2024 \\ \textbf{DSv1.1}} & \thead{2025 \\ \textbf{DSv2}}  &\thead{2025  \\ \textbf{CDFv3}}& \textbf{Mean} \\
        \midrule
ForAda~\cite{ForensicsAdapter} & \textbf{93.9} & \textbf{77.1} & 31.4 & 28.6 & \textbf{55.3} & 31.4 & \textbf{67.2} & 59.7 & 75.8 & 29.5 & 45.9 & 30.9 & 6.5 & 18.2 & 46.5 \\
Effort~\cite{Effort} & 89.8 & 75.8 & \textbf{44.7} & 15.7 & 47.1 & \textbf{55.8} & 54.6 & 61.5 & 79.7 & 20.5 & 64.8 & \textbf{34.7} & 3.8 & 27.9 & 48.3 \\
GenD (CLIP) & 92.9$\pm$1.4 & 72.3$\pm$2.0 & 30.4$\pm$0.1 & \textbf{34.6$\pm$5.6} & 37.9$\pm$10.4 & 38.0$\pm$2.6 & 60.7$\pm$0.7 & 66.9$\pm$0.9 & 92.3$\pm$0.2 & 19.2$\pm$4.4 & 78.8$\pm$3.0 & 28.4$\pm$0.5 & 10.9$\pm$4.0 & 23.8$\pm$0.7 & 49.1 \\
GenD (PE) & 86.1$\pm$5.8 & 74.6$\pm$4.4 & 17.6$\pm$2.3 & 23.9$\pm$5.0 & 33.2$\pm$7.8 & 29.9$\pm$5.8 & 62.7$\pm$1.7 & 72.0$\pm$5.6 & 78.3$\pm$2.5 & 26.8$\pm$2.5 & 79.4$\pm$2.9 & 16.0$\pm$3.8 & \textbf{29.6$\pm$3.1} & \textbf{30.4$\pm$3.9} & 47.2 \\
GenD (DINO) & 82.0$\pm$6.0 & 69.8$\pm$4.3 & 34.8$\pm$3.2 & 31.4$\pm$7.9 & 21.8$\pm$5.8 & 41.9$\pm$7.6 & 66.1$\pm$1.2 & \textbf{76.2$\pm$7.2} & \textbf{96.3$\pm$1.6} & \textbf{33.6$\pm$4.3} & \textbf{82.0$\pm$7.8} & 25.5$\pm$2.7 & 14.6$\pm$5.3 & 16.4$\pm$5.4 & \textbf{49.5} \\
    \bottomrule
    \end{tabular}%
    }
\end{table*}

\begin{table*}[t]
    \centering
    \caption{\label{tab:ablation-full}
    Ablation study of L2 normalization and uniformity-alignment (UA) without LN-tuning. LP is the linear probing.}

    \resizebox{\textwidth}{!}{%
    \begin{tabular}{l|cccccccccccccc|c}
        \toprule
         \textbf{Method} & \thead{2019 \\ \textbf{UADFV}} & \thead{2019 \\ \textbf{DFD}} & \thead{2019 \\ \textbf{DFDC}} & \thead{2020 \\ \textbf{FSh}} & \thead{2020 \\ \textbf{CDFv2}} & \thead{2021 \\ \textbf{FFIW}} & \thead{2021 \\ \textbf{KoDF}} & \thead{2021 \\ \textbf{FAVC}} & \thead{2022 \\ \textbf{DFDM}} & \thead{2024 \\ \textbf{PGF}} & \thead{2024 \\ \textbf{IDF}} & \thead{2024 \\ \textbf{DSv1.1}} & \thead{2025 \\ \textbf{DSv2}}  &\thead{2025  \\ \textbf{CDFv3}}& \textbf{Mean} \\
        \midrule
CLIP+LP & 94.6$\pm$0.7 & 89.2$\pm$1.2 & \textbf{75.3$\pm$0.3} & \textbf{77.6$\pm$0.9} & 74.6$\pm$1.2 & 80.7$\pm$1.9 & \textbf{81.8$\pm$1.3} & \textbf{83.0$\pm$1.3} & 77.8$\pm$1.6 & 62.7$\pm$0.4 & 68.7$\pm$3.9 & \textbf{64.5$\pm$1.0} & \textbf{57.8$\pm$1.0} & 75.6$\pm$1.3 & 76.0 \\
CLIP+LP+L2 & \textbf{97.3$\pm$1.3} & \textbf{89.2$\pm$0.7} & 73.7$\pm$1.9 & 77.4$\pm$2.3 & \textbf{76.5$\pm$0.7} & \textbf{84.2$\pm$0.6} & 77.1$\pm$3.0 & 80.6$\pm$1.5 & \textbf{84.7$\pm$2.2} & \textbf{68.1$\pm$0.5} & \textbf{84.0$\pm$5.4} & 62.3$\pm$3.2 & 57.1$\pm$0.9 & \textbf{77.2$\pm$0.7} & \textbf{77.8} \\
CLIP+LP+L2+UA & \textbf{97.3$\pm$1.3} & \textbf{89.2$\pm$0.7} & 73.7$\pm$1.9 & 77.4$\pm$2.3 & \textbf{76.5$\pm$0.7} & \textbf{84.2$\pm$0.6} & 77.1$\pm$3.0 & 80.6$\pm$1.5 & \textbf{84.7$\pm$2.2} & \textbf{68.1$\pm$0.5} & \textbf{84.0$\pm$5.4} & 62.3$\pm$3.2 & 57.1$\pm$0.9 & \textbf{77.2$\pm$0.7} & \textbf{77.8} \\
        \bottomrule
    \end{tabular}%
    }
\end{table*}

\end{document}